\documentclass[twoside]{article}
\usepackage{subcaption}
\usepackage{amssymb,amsmath, bm, color, bbm}
\usepackage{graphicx}
\usepackage[normalem]{ulem}
\usepackage[]{algorithm2e}
\usepackage{url}
\usepackage{hyperref}
\newtheorem{theorem}{Theorem}
\newtheorem{corollary}{Corollary}
\newtheorem{definition}{Definition}
\newtheorem{proof}{Proof}

\begin{document}

\title{A Minimax Surrogate Loss Approach to Conditional Difference Estimation}
\author{Siong Thye Goh \\ MIT \and Cynthia Rudin \\ Duke University}
%\aistatsauthor{ Siong Thye Goh \And Cynthia Rudin}

%\aistatsaddress{ Massachusetts Institute of Technology \And  Duke University} 
\maketitle

\begin{abstract}
We present a new machine learning approach to estimate personalized treatment effects in the classical potential outcomes framework with binary outcomes. %We present a new 0-1 loss function.
To overcome the problem that both treatment and control outcomes for the same unit are required for supervised learning, we propose surrogate loss functions that incorporate both treatment and control data. The new surrogates yield tighter bounds than the sum of losses for treatment and control groups. A specific choice of loss function, namely a type of hinge loss, yields a minimax support vector machine formulation. The resulting optimization problem requires the solution to only a single convex optimization problem, incorporating both treatment and control units, and it enables the kernel trick to be used to handle nonlinear (also non-parametric) estimation. Statistical learning bounds are also presented for the framework, and experimental results.
%Our simulation results show that our approach are compatible with existing algorithms.
\end{abstract}

\section{Introduction}

Many data-driven decisions, such as whether to prescribe a particular pharmaceutical drug or whether to launch a particular marketing campaign, are problems of causal inference that require conditional difference estimation. Causal inference considers the effects of interventions, which is the basis for policy-making. It is well-known that standard machine learning methods are not designed to handle questions of causal inference; they are designed only for prediction and not for estimation of conditional differences or causal effects. A key reason that supervised machine learning does not usually handle causal inference problems is that by the nature of these problems, we do not observe counterfactuals (e.g., what would have happened if the unit had not received the treatment), which means we are missing half of each label of a supervised learning problem. On the other hand, machine learning can handle powerful nonlinear modeling problems, which traditional causal inference methods cannot. Ideally, we would leverage the strengths of modern machine learning to create powerful models for conditional differences that could, in the right settings, be used for causal inference.
 
This work provides an approach to nonlinear treatment effect estimation using machine learning, where the outcomes are binary (yes/no), and the goal is to predict whether treatment effects are positive, neutral, or negative. Given a new sample of units that are not in the training set, our goal is to decide which units would have a positive treatment effect, and which units would have a negative treatment effect. In this setting, we assume we can personalize who receives the treatment. Since the treatment is not globally launched across the population, it does not make sense to investigate average treatment effects. Since the outcome is binary, we are not interested in the estimated size of the treatment effect, but the simpler question of whether we have correctly determined whether the treatment effect is positive or negative for each individual. This is related to policy questions such as what fraction of the population did we correctly assigned the treatment. 

We present a single formulation that handles treatment group and control group data simultaneously, and outputs a single function $f$ whose thresholds at $-1$ and 1 provide decision boundaries between positive, neutral, and negative treatment effects. We provide a formulation as a type of ``minimax'' support vector machine. This handles either linear or non-linear treatment responses in a computationally efficient manner (via the kernel trick). By changing the kernel we create nonparametric models, and if we use the natural linear kernel, we create linear decision boundaries like regression models. 
 
A large body of work in the causal inference community has focused on estimating average treatment effects (ATE) through linear models \cite{rubin1974estimating, rubin1977assignment, rubin1981estimation,holland1986statistics, robins1995analysis, scharfstein1999adjusting, bang2005doubly, belloni2014high}, where the coefficient for the treatment variable provides an estimate for the ATE. As discussed earlier, ATE estimation is not relevant for determining who should receive a (personalized) treatment.
%\citep{angrist1995identification,heckman2001instrumental,wooldridge1997two,imbens2004nonparametric,wooldridge2008instrumental,angrist1991instrumental}. Conversely, machine learning focuses on personalized predictions, and is not focused on ATE estimation, leading to nonlinear models.
%For personalized prediction problem, the underlying structure could be nonlinear. It is natural to use machine learning to handle such problems as machine learning has illustrated that it can handle nonlinear decision boundaries, which has been helpful for SVM, boosting atc.
Recently there has been a lot of work on subgroup identification \cite{imai2010identification, imai2013experimental,ratkovic2017sparse}
% WangRu15CFRL} 
and other types of nonlinear predictors of causal effects \cite{athey2015machine, johansson2016learning,athey2016recursive,wager2017estimation,2016arXiv160603976S}. %In \citep{}, under the ignorability assumption, a neural network approach is proposed to learn nonlinear treatment effects. 
Our work also focuses on personalized predictions of treatment effects, but differs from those listed above in several ways: 
% \begin{itemize}
%\item 

1) Our methods construct a single model with a tighter bound on the surrogate loss than a sum of treatment and control losses. This means that theoretically, our method should create more accurate single models than if one created separate treatment and control models and subtracted them to obtain an estimate for the treatment effect. This arises from our formulation as a single regularized minimax problem. 
%\item 
%With a linear kernel, it becomes a single linear regression. With a radial basis function (RBF) kernel, it becomes a nonparametric model. Polynomial kernels are a graceful middle ground between linear regression models and RBF kernels.

2) The algorithm does not rely on greedy splitting and pruning heuristics or other non-convex optimization procedures, such as decision trees, random forests, matching, neural networks, etc. Our formulation is a single convex quadratic optimization problem that has known fast solution methods. Model complexity depends on the choice of kernel and regularization parameters, not on splitting or pruning parameters.
%\item The support vector machine formulation creates support vectors, which lie on the boundary between classes, and the solution is expressed in terms of them.
%\end{itemize}

One work that seems similar to ours on the surface but is not, is that of Ratkovic and Tingley \cite{Ratkovic2014BalancingWT}, who use support vector machines (SVM's) only to determine the largest balanced subset of data, by classifying which units are likely to have high density according to the treatment population distribution. From there, a traditional method is used to estimate conditional differences. Conversely, in our work, we use a traditional inverse propensity score model \cite{sugiyama2008direct,hido2011statistical} or other method to estimate the ratio of densities, and propose a single support vector machine formulation to estimate treatment effects. 

Another work that is more relevant to ours is that of \cite{imai2013} who use a regularized squared hinge loss over all observations to estimate a single model that predicts outcomes for both treatment and control. The model includes two sets of covariates to predict outcomes, one that does not depend on the treatment and the other that does. 
%For predicting conditional differences, the second set of covariates would not be used. 
Usually hinge loss is chosen to be a convex proxy for a given 0-1 loss that is hard to minimize, but it is not clear what the 0-1 loss is in their case, as it is not discussed. The 0-1 loss implicit in their formulation seems to be the sum of 0-1 losses for predictions of both treatment and control \textit{outcomes} rather than a 0-1 loss for \textit{treatment effects}. In this work, we prove that the 0-1 loss for prediction of outcomes is an upper bound on a relevant 0-1 loss for treatment effects, motivating the use of their 0-1 loss, but showing that it is a loose upper bound. In our formulation we use a tighter upper bound than the sum of 0-1 losses for prediction. The method of \cite{imai2013} is similar to estimating treatment and control outcomes with two separate models, because the estimated control outcomes can depend on one set of covariates (forming one model), whereas the treatment outcomes can depend on the other (treatment-related) set of covariates (forming the second model). \footnote{In fact their method would be identical to the two-separate-models approach if their model is chosen to be an indicator for control times a linear combination of variables plus an indicator for treatment times a linear combination of variables.}

Let us discuss the estimation of density ratios in order to perform inverse propensity score weighting. We would use inverse propensity score weighting to correct for the fact that the control and/or treatment data does not come directly from the target population of interest.
In general, if the (conditional) treatment effects are estimated correctly, then it is irrelevant whether or not the density ratios are poorly estimated. As an extreme case, if both the treatment and control losses are zero, the density ratio estimate is completely irrelevant. Thus, if we focus directly on accurate estimates of treatment effects, we may avoid problems faced by other methods that use looser surrogate loss functions. 
%Our surrogate loss is strictly tighter than a sum of separate losses from treatment and control groups.

In cases where conditional treatment effects are not able to be perfectly estimated, our method still can provide high quality treatment effect estimates without accurate estimation of the density ratio. Our predictor function minimizes the larger of the treatment loss and the control loss. If the target population is the treatment population, then the control loss involves density ratio estimation but not the treatment loss. Hence, if the treatment loss is always higher than the re-weighted control loss, then regardless of whether the density ratio is poorly estimated, the method will still produce the same answer. Its result is robust to poor density estimates when this happens.

%this is the motivation for our approach.
 
Because support vector machines with radial basis functions are nonparametric, they are related to matching approaches. Historically, matching methods \cite{rosenbaum1983assessing,rosenbaum1984consquences, dehejia2002propensity,zubizarreta2012using, keele2015enhancing,king2017balance} are different in that the matching is done prior to the modeling, with some exceptions \cite{Tianyu2017}; here we use a single modeling approach.

   \section{Problem Setting}\label{sec:probsetting}

We work in a standard potential outcomes setting, with observational data. Each observation possesses covariates %(assuming no unmeasured confounding) 
and is assigned to either treatment or control groups,  and an outcome is observed for each individual. 
The potential outcomes for observation $i$ are denoted by $Y_i^T$ or $Y_i^C$, where the superscript $T$ denotes membership in the treatment group, and $C$ denotes the control group. We cannot observe instances of $Y_i^T-Y_i^C$ since $Y_i^T$ and $Y_i^C$ are not simultaneously observable. Hence, this is a missing data problem where exactly half of the data are not observable. We define a binary causal exposure variable $W$, taking value 1 if the corresponding sample point belongs to the treatment group and 0 for control points. The outcome variable $Y_i^{obs}$ is thus: $$Y_i^{obs}=WY_i^T+(1-W)Y_i^C.$$  

%Causal effects are defined in rows of Table \ref{fundamental} \citep{holland1986statistics}.
%\begin{table}[htbp]
%\begin{center}
%  \begin{tabular}{ |c | c |  c| }
%    \hline
%    Group & $Y^T$ & $Y^C$ \\ \hline
%    Treatment ($W=1$) & Observable & Counterfactual \\ \hline
%    Control ($W=0$)& Counterfactual & Observable \\
%    \hline
%  \end{tabular}\caption{The fundamental problem of causal inference}
%  \label{fundamental}
%\end{center}
%\end{table}

We assume outcomes $Y_i^T$  and $Y_i^C$ depend on features (covariates) of the data. The features follow distributions $\mu_{X|T}$ and $\mu_{X|C}$ for treatment and control, respectively. Covariates are denoted by $X$. We often use upper case for random variables and lower case for draws of random variables. In notation, $Y^T \sim \mu_{Y^T|x}$, and $Y^C \sim \mu_{Y^C|x}$. Since we assume underlying treatment and control populations differ on the covariate space, they are notated as $\mu_{X|T}$ and $\mu_{X|C}$, where treatment observations follow $X\sim \mu_{X|T}$, and control observations follow $X\sim \mu_{X|C}$. By using the standard Radon-Nikodym derivative, we can effectively transform one distribution to another. The method we introduce can be trivially adapted to any target distribution, so for ease of notation we chose the treatment distribution $\mu_{X|T}$ to be our target population. 

In this paper, we consider binary (yes/no) outcomes $y^T,y^C \in \left\{ -1,1 \right\}$, e.g., whether or not someone had a heart attack. We let $h$ denote our predictor function of $Y^T-Y^C$, which is a function of the covariates.

If we were given a predictor function \textit{and the ground truth}, we might measure the quality of our predictor function using the following two conditional-difference loss functions.
The first one is:
$$l_{0-1}(x,y^T,y^C, h)=\begin{cases} \mathbbm{1}_{|h(x)| \geq 1}, & y^T=y^C  \\
\mathbbm{1}_{h(x)\leq 0 }, &  y^T>y^C \\
\mathbbm{1}_{h(x) \geq 0}, &  y^T<y^C.
\\\end{cases}.$$
This loss is 1 if there is no treatment effect and $h$ predicts either a positive or negative treatment effect (top condition). The loss is 1 also when $h$ predicts a treatment effect that is opposite from the true treatment effect. This loss function does not consider the average or magnitude of the treatment effect, it counts the number of individuals for whom the treatment effect was incorrectly predicted. This is a relevant loss when we aim to correctly assign treatment to individual members of a population: e.g., optimally assigning advertisements to individuals visiting a website, or optimally assigning a pharmaceutical drug to individuals who would benefit from it.

The second loss function that we consider is
$$l_{\theta}(x,y^T,y^C, h)=\begin{cases} \mathbbm{1}_{|h(x)|\geq \theta}, & y^T=y^C  \\
\mathbbm{1}_{h(x)\leq -\theta}, &  y^T>y^C \\
\mathbbm{1}_{h(x) \geq \theta}, &  y^T<y^C.
\\\end{cases}$$
The first loss function is an upper bound for the second loss function, for margin $\theta>1$. 
For the second loss function, $l_{\theta}$, the set of possible inputs $x$ for which $h(x)$ is within $(-\theta, \theta)$ can be interpreted as the region with no large predicted treatment effect. Provided that the predictions of treatment effect are real-valued, $h(x)$ can be rescaled, and thus to minimize $l_{\theta}$ it suffices to consider $l_{1}$ by letting $\theta = 1$.

$$l_{1}(x,y^T,y^C, h)=\begin{cases} \mathbbm{1}_{|h(x)| \geq 1}, & y^T=y^C  \\
\mathbbm{1}_{h(x)\leq -1}, &  y^T>y^C , \text{(false negative)}\\
\mathbbm{1}_{h(x)\geq 1}, &  y^T<y^C , \text{(false positive)}.
\\\end{cases}$$

To see this, note that suppose we have $h_1 \in \arg \min_{h} l_1(.) $, then $\theta h_1 \in \arg \min_h l_{\theta}(.).$ Hence, we will focus on $l_{\theta}$ where $\theta=1$  when we derive our algorithm.

Since $y^T$ and $y^C$ are not observed simultaneously, it is not possible to compute the quantity above. This motivated us to use a surrogate function that separates $y^T$ and $y^C$. Ideally, a surrogate function is an upper bound to the 0-1 loss, with a minimizer that can be easily computed. We dedicate the next section to an upper bound of the conditional-difference 0-1 loss function which in turn motivates the surrogate function explored in Section \ref{sec:framework}. 

\section{A Surrogate Conditional-Difference Loss Function \label{sec:Uppbound01}}

The following theorem defines sufficient conditions under which a surrogate loss function is  valid for $l_1$. 
\begin{theorem}\label{theoremupperbd}
If a function $l(.)$ satisfies $l(z) \geq \mathbbm{1}_{z \geq 0} + \mathbbm{1}_{z \geq 1}$, then we have  
\begin{align*}
&\mathbb{E}_{X \sim \mu_{X|T},Y^T \sim \mu_{Y^T|X},Y^C\sim \mu_{Y^C|X}}l_{1}(X,Y^T,Y^C, h) 
\\ &\leq \max \left(  \mathbb{E}_{X \sim \mu_{X|T}, Y^T \sim \mu_{Y^T|X}} l(-h(X)Y^T),  \mathbb{E}_{X \sim \mu_{X|C} , Y^C \sim \mu_{Y^C|X}} \frac{ l(h(X)Y^C) }{\mu_{X|C}(X)/\mu_{X|T}(X)} \right).
\end{align*}
\end{theorem}
The proof of the theorem is in the appendix. It involves finding a lower bound for each loss term in the maximum function, symmetry arguments, and using properties of indicator variables. It is broken down in $5$ subsections in the proof to facilitate the reader's understanding.

We have a similar bound for the $l_{0-1}$ loss function,

\begin{theorem}\label{theoremupperbdzero}
If a function $l(.)$ satisfies $l(z) \geq \mathbbm{1}_{z \geq 0} + \mathbbm{1}_{z \geq 1}$, then we have  
\begin{align*}
&\mathbb{E}_{X \sim \mu_{X|T},Y^T \sim \mu_{Y^T|X},Y^C\sim \mu_{Y^C|X}}l_{0-1}(X,Y^T,Y^C, h) \\ &\leq \max \left(  \mathbb{E}_{X \sim \mu_{X|T},Y^T \sim \mu_{Y^T|X}} l(-h(X)Y^T), \mathbb{E}_{X \sim \mu_{X|C} ,Y^C \sim \mu_{Y^C|X}} \frac{ l(h(X)Y^C) }{\mu_{X|C}(X)/\mu_{X|T}(X)}\right).
\end{align*}
\end{theorem}
The proof of this theorem in \cite{ARXIV} is similar to that of Theorem \ref{theoremupperbd}.

The subscript of the expectation includes the generative model for the data. Here, $\mathbb{E}_{X \sim \mu_X,Y^T \sim \mu_{Y^T|X},Y^C\sim \mu_{Y^C|X}}$  means that, first, $X$ follows distribution $\mu_X$, then $Y^T$ is drawn from distribution $\mu_{Y^T|X}$, and $Y^C$ is drawn from distribution $\mu_{Y^C|X}$.

The significance of this inequality is that the quantity on the right can be estimated using empirical averages, without imputation for counterfactuals. 
%The proof of this result is provided in the supplementary document. 

The right-hand sides of the theorems are our surrogate loss functions. Since the max of the two terms is less than or equal to their sum, our surrogate losses are strictly tighter than using the sum of treatment and control losses. That sum would lead to separate modeling for treatment and control groups.

One corollary of the theorem is a remark on the importance of accurate density ratio estimation. The following corollary shows that in some cases, it is not crucial to obtain an accurate estimate of the density ratio.

\begin{corollary}
If for all functions $h$,
\begin{align*}
R(h) :=&\max \left(  \mathbb{E}_{X \sim \mu_{X|T},Y^T \sim \mu_{Y^T|X}} l(-h(X)Y^T),  \mathbb{E}_{X \sim \mu_{X|C} ,Y^C \sim \mu_{Y^C|X}} \frac{ l(h(X)Y^C) }{\mu_{X|}(X)/\mu_{X|T}(X)}\right) \\&=  \mathbb{E}_{X \sim \mu_{X|T},Y^T \sim \mu_{Y^T|X}} l(-h(X)Y^T), \textrm{   and also}
\end{align*}
\begin{align*}
\hat{R}(h):= &\max \left(  \mathbb{E}_{X \sim \mu_{X|T},Y^T \sim \mu_{Y^T|X}} l(-h(X)Y^T), \mathbb{E}_{X \sim \mu_{X|C} ,Y^C \sim \mu_{Y^C|X}} \frac{ l(h(X)Y^C) }{\hat{\mu}_{X|C}(X)/\hat{\mu}_{X|T}(X)}\right) \\&=  \mathbb{E}_{X \sim \mu_{X|T},Y^T \sim \mu_{Y^T|X}} l(-h(X)Y^T),
\end{align*} 
then the minimizers $\min_h R(h)$ and $\min_h \hat{R}(h)$ do not depend on how close the estimates $\hat{\mu}_{X|C}(X)/\hat{\mu}_{X|T}(X)$ are to the true ratios $\mu_{X|C}(X)/\mu_{X|T}(X)$.
\end{corollary}

\textcolor{blue}{}

The corollary implies that flaws in density ratio estimation have no effect in a special case where the treatment loss is higher than the re-weighted control loss. This is different from using the sum of losses for treatment and control groups, where problems with density ratio estimation can affect the loss regardless of which achieves the max. Thus, our method is sometimes robust to poor density estimation methods, and where it is not, the same problem is present in traditional methods; our method is no worse. 

%Intuitively, in cases where predictions are close to perfect, the loss is small and poorly estimated weights cannot change the predictions very much.

%%%%%%%%%%%%%%%%%%%%%%

%section{Example of Functions which satisfy the condition to construct surrogate functions}

The sufficient condition to construct a surrogate upper bound for the 0-1 loss function is used in both theorems. It is:  
$$l(z) \geq \mathbbm{1}_{z \geq 0} + \mathbbm{1}_{z \geq 1}.$$
This condition is easy to satisfy, and we list some valid losses below. 
\begin{enumerate}
\item $\mathbbm{1}_{z \geq 0} + \mathbbm{1}_{z \geq 1}$, which is a type of 0-1 loss function. Clearly, this is trivially an upper bound, and it is non-smooth and it is difficult to optimize directly.
\item $\lfloor 1+z \rfloor_+$, the hinge loss function. We will use this loss function to construct an SVM-based algorithm.
\item $(1+z)^2$, the squared loss function. 
\item $\frac{2\ln (1+e^z)}{\ln(1+e)}$, a scaled logistic loss function.
\item $e^z$, the exponential loss, used by AdaBoost.
\end{enumerate}

%In the next section, we formulate our SVM framework using the hinge loss function.

\section{Conditional Difference SVM \label{sec:framework}}

In this section, we use the regularized hinge loss to formulate a quadratic programming problem that is similar to classical SVM (except that it is for potential outcomes data where we have only ``half" of the label for each observation).

For this section, we assume that the ratio $\mu_C(x_i)/\mu_T(x_i)$ is either known or has been estimated previously (or is irrelevant according to the theorems above). We will discuss this more later. These density ratios act as importance weights on the control group terms for \textit{cost-sensitive} learning. 
%While weights like these do influence the quality of classification, performance may not be especially sensitive to them. 
%\textcolor{red}{Consider, for instance, the double spiral example in Section \ref{sec:probsetting}. 
%A change of weights cannot change the predictions very much since they are close to perfect. On the other hand, the choice of single SVM versus the ``double" SVM formulation makes a huge difference, as we showed. }
The formulation below is kernelized, meaning that each $x$ is replaced with a transformation $\phi(x)$, such that $\langle \phi(p), \phi(q) \rangle = K(p,q)$ can be evaluated efficiently as a kernel function. There are several standard conditions for $k$ to be a valid kernel (an inner product of a Reproducing Kernel Hilbert space -- RKHS). Trivially, if $\phi$ is chosen to be the identity, then the kernel is linear. The model for $h(x)$ is $w_0+\langle \phi(w), \phi(x) \rangle$.

The optimization problem suggested by Theorem \ref{theoremupperbd}, with the hinge loss as an upper bound on the 0-1 loss, is below. We added an RKHS norm regularization term with regularization parameter $\gamma$.
\begin{eqnarray*}
\lefteqn{R(w,w_0,\gamma)}\\
&=\max\left( \frac1{n^T}\sum_{i \in T} \lfloor 1-(w_0+\langle \phi(w), \phi(x_i) \rangle)y_i^T \rfloor_+, \right. \\& \left. \frac1{n^C}\ \sum_{i \in C}  \frac{\lfloor 1+(w_0+\langle \phi(w), \phi(x_i) \rangle)y_i^C  \rfloor_+}{\mu_{X|C}(x_i) / \mu_{X|T}(x_i)}\right)\\
&+\gamma \langle  \phi(w), \phi(w) \rangle.
\end{eqnarray*}
Rewriting the inner product as a kernel, this is equivalent to:
\begin{eqnarray*}
\lefteqn{R(w,w_0,\gamma)}\\
&=\max\left( \frac1{n^T}\sum_{i \in T} 
\left\lfloor 1-(w_0+ K(w,x_i) )y_i^T \right\rfloor_+, \right. \\&\left.\frac1{n^C}\ \sum_{i \in C}  \frac{
\left\lfloor 1+(w_0+ K(w,x_i) )y_i^C\right\rfloor_+}{\mu_{X|C}(x_i) / \mu_{X|T}(x_i)}  \right)+\gamma K(w,w).
\end{eqnarray*} 
This minimax problem can be reformulated as a constrained optimization problem as follows:

\noindent \textbf{Primal Problem:}
\begin{eqnarray*}
\lefteqn{\min_{w,w_0,z,r,\forall i \;s_i,\forall i \; r_i} z+ \gamma K(w,w) \;\;\;\textrm{ subject to}}\\
z &\geq & \frac{1}{n^T}\sum_{i \in T} r_i\\
z &\geq & \frac{1}{n^C}\sum_{i \in C} \frac{s_i}{\mu_{X|C}(x_i)/\mu_{X|T}(x_i)}\\
r_i &\geq & 1-(w_0+K(w,x_i))y_i^T, \forall i \in T\\
s_i &\geq & 1+(w_0+K(w, x_i))y_i^C, \forall i \in C\\ 
r_i &\geq & 0, \forall i \in T\\
s_i &\geq & 0, \forall i \in C.
\end{eqnarray*}

We let $K^*$ be the Gram matrix where $K^*(i,j)=K(x_i,x_j)$ where we order the vectors such that $x_1^T, \ldots, x_{n_T}^T, x_1^C, \ldots, x_{n_C}^C$. %Focusing on terms that involve $K(.,.)$ in the Lagrangian, we simplify the optimization problem by replacing the primal variables by functions of the dual variables using the KKT conditions (\ref{alphaplusbeta1})-(\ref{diffsi}) that we obtained earlier. 
%Part of the Lagrangian reduces as follows: 
%\begin{align*}
%&\gamma K(w,w) + \sum_{i \in T} \lambda_i (-K(w,x_i)y_i^T) + \sum_{i \in C} \eta_i (K(w,x_i)y_i^C) \\
%&= -\frac{1}{4\gamma}\left[\begin{array}{cc} \lambda^T & \eta^T\end{array} \right]diag(y_1^T, \ldots, y_{n^T}^T, -y_1^C, \ldots, -y_{n^C}^C)\\&K^* diag(y_1^T, \ldots, y_{n^T}^T, -y_1^C, \ldots, -y_{n^C}^C)\left[ \begin{array}{c}  \lambda \\ \eta \end{array}\right]. 
%\end{align*}
%The first equality above follows from (\ref{dualconstraint}) and from writing the sums as a vector product. The second equality above follows from simplifying, and from using (\ref{dualconstraint}) again. 
%Hence, 

The corresponding dual optimization problem is as follows.

\noindent \textbf{Dual Problem:}
\begin{align*}
&\max_{\alpha,\beta,\{\lambda_i\}_i,\{\eta_i\}_i} -\frac{1}{4\gamma}
\left[\begin{array}{c} \lambda \\ \eta\end{array} \right]^T 
diag(y_1^T, \ldots, y_{n^T}^T, -y_1^C, \ldots, -y_{n^C}^C)K^* \\&diag(y_1^T, \ldots, y_{n^T}^T, -y_1^C, \ldots, -y_{n^C}^C)\left[ \begin{array}{c}  \lambda \\ \eta \end{array}\right]+\sum_{i \in T} \lambda_i+ \sum_{i \in C} \eta_i,
\end{align*}
subject to
\begin{eqnarray*}
\alpha+\beta&=1\\
\forall i \in T, &0 \leq \lambda_i \leq \frac{1}{n^T} \alpha \\
\forall i \in C, &0 \leq \eta_i \leq \frac{1}{n^C(\mu_{X|C}(x_i)/\mu_{X|T}(x_i))}\beta \\
\lefteqn{\sum_{i \in T}\lambda_iy_i^T = \sum_{i \in C} \eta_i y_i^C}
\\
\alpha, \beta, \lambda, \eta &\geq 0
\end{eqnarray*}
which is a quadratic programming problem that resembles the regular SVM problem. Its computational scaling properties are essentially identical to standard SVM.

\noindent \textbf{Recovering the Intercept $w_0$}

\begin{figure}[h]
    \centering
    \includegraphics[height=0.8in]{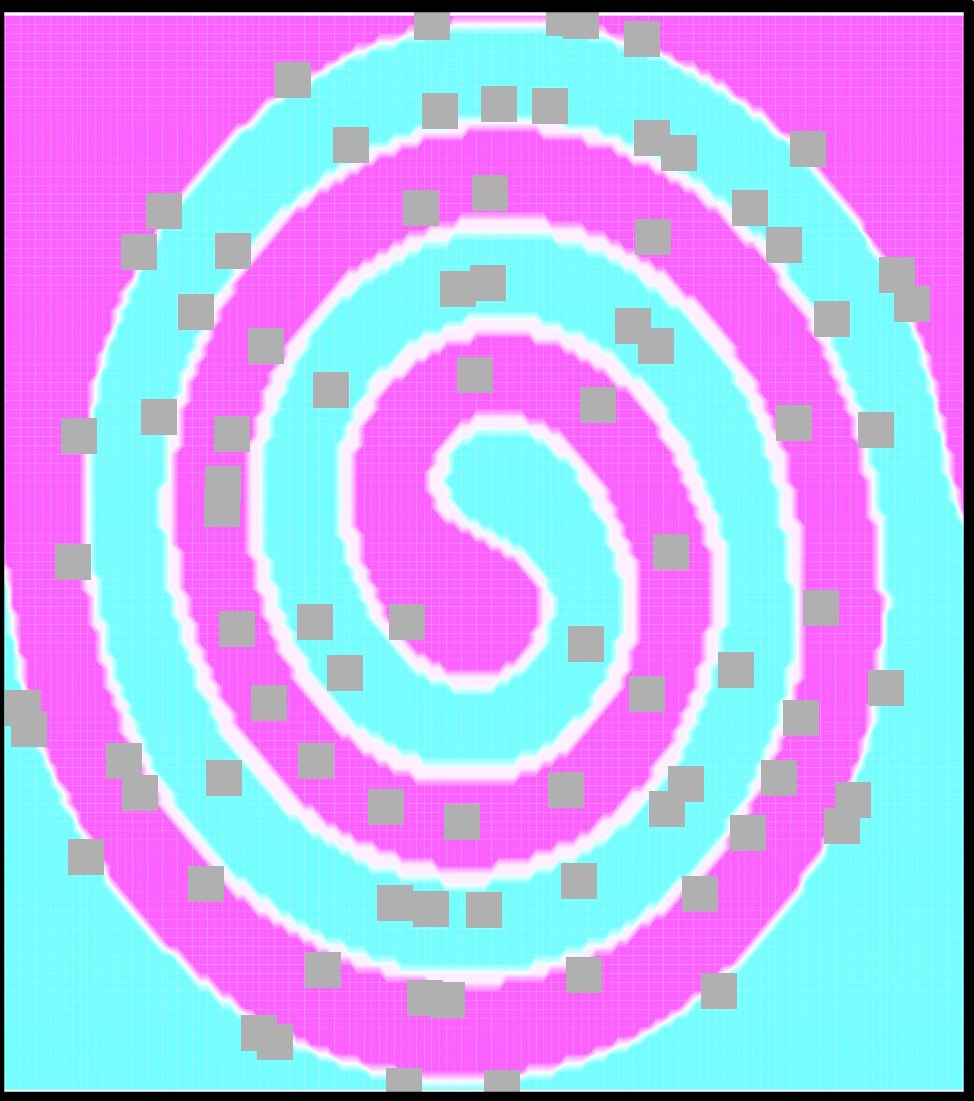}
    \caption{Causal SVM with RBF kernel on spiral data, the circular points are the support vectors, pink indicates predictions of positive treatment effect, and blue indicates negative predictions.\label{Fig:sv}}
 \end{figure}

After solving for $\lambda$ and $\eta$, we are able to theoretically recover an expression for $\phi(w)$ in the primal formulation that can be used to obtain values of $K(w,x)$ for any given $x$.
To make prediction possible, we need to evaluate $h(x)$ for any $x$, thus we need to recover $w_0$, the intercept term.
The complementary slackness conditions are as follows:
\begin{eqnarray*}
\lambda_i (r_i-1+(w_0+K(w,x_i^T))y_i^T)=0\;\; \forall i\in T\\
\eta_i (s_i-1-(w_0+K(w,x_i^C))y_i^C)=0\;\; \forall i \in C\\
r_i \left( \frac{\alpha}{n^T}-\lambda_i\right)=0 \;\; \forall i\in T\\
s_i \left( \frac{\beta}{n^C(\mu_{X|C}(x_i)/\mu_{X|T}(x_i))}-\eta_i\right)=0 \;\; \forall i \in C.
\end{eqnarray*}
By solving the dual optimization problem, we know the value of  $\{\lambda_i\}_i, \{\eta_i\}_i, \alpha, \beta$. We can use these to analytically recover $w_0$ from the primal problem using one of the ``support vectors." Support vectors are data points that determine the separating hyperplane in regular SVM. 
 In our context, we are minimizing the maximum of two hinge losses, and the maximum value will be attained by at least one of the control or treatment group. The hyperplane is chosen such that it minimizes loss (and maximizes the margin) in one of those groups, and since the larger of the two losses is being minimized, the loss in the other group will be upper bounded as well. 
%Not every data point will influence the position of the separating hyperplane. 
Similar to the regular SVM, the points that fully determine the positions of the hyperplane (support vectors - SV's) are those with active constraints in the primal formulation. Figure \ref{Fig:sv} shows the support vectors for the Causal SVM on the spiral dataset discussed below. SV's from both treatment and control points can be present simultaneously.
%\textcolor{red}{be more precise - When the condition is regular enough}. 
As usual, as long as the problem is not ill-conditioned (meaning at least one $\lambda_i$ is between 0 and $\alpha/n^T$, or at least one $\eta_i$ is between 0 and $\beta/n^C$), we are able to recover the primal solution from the dual solution as follows: for $i\in T$ if $\lambda_i < \frac{\alpha}{n^T}$, we can conclude that $r_i=0$ and similarly if $\lambda_i >0$, we can conclude that $-1+(w_0+K(w,x_i))y_i^T=0$.
Using that $y_i$ is binary: 
$$w_0=y_i^T-K(w,x_i^T).$$
Similarly, for $i\in C$ if $\eta_i < \frac{\beta}{n^C(\mu_{X|C}(x_i)/\mu_{X|T}(x_i))}$, we conclude that $s_i=0$, and if for the same $i \in C$, $\eta_i>0$, we have 
$w_0=-y_i^C-K(w,x_i^C).$
Also,  using optimization methods that use a primal dual approach, it is possible to obtain $w_0$ numerically.

Let us switch gears to discuss learning theory bounds.

\section{Generalization Bound}
The bound in this section provides a theoretical foundation for minimizing the maximum of treatment and control empirical errors. 
%The bound in this section provides a foundation for using the minimax loss provided above.
%We first introduce some terminology.
% in order to state a statistical learning bound for causal inference. 

\begin{definition}{Growth Function: \cite{bousquet2004introduction} }{ Let $\mathcal{F}$ be a function class (also known as hypothesis class). Given data points $z_1, \ldots, z_m$, we consider $\mathcal{F}_{z_1,\ldots, z_m}=\{ f(z_1), \ldots, f(z_m) \}$, the set of ways the data $z_1, \ldots, z_m$ are classified by functions form $\mathcal{F}$. The growth function is the maximum number of ways into which $m$ points can be classified by the function class. $S_{\mathcal{F}}(m) = \sup_{(z_1, \ldots, z_m)}|\mathcal{F}_{z_1, \ldots, z_m}|.$ } \end{definition}

Let $R^{true}(f)=\mathbb{P}_{(x,y) \sim D}(f(X) \neq Y)=\mathbb{E}_{(X,Y) \sim D} [\mathbbm{1}_{f(X) \neq Y}]$ and $R^{emp}(f) = \frac{1}{m}\sum_{i=1}^m \mathbbm{1}_{[f(x_i) \neq y_i]}$. Using the Hoeffding and union bounds, a classical result shows that for any $\delta > 0$, with probability at least $1-\delta$ with respect to a random draw of the data, 
$$\forall f \in \mathcal{F}, R^{true}(f) \leq R^{emp}(f) + 2 \sqrt{2\frac{\log S_{\mathcal{F}}(2m) + \log \frac{4}{\delta}}{m}}.$$
We will derive an analogous bound for the causal inference estimation framework. Since we deal with both treatment and control groups, we need to handle weighted data points with Radon-Nikodym derivatives. More definitions follow.

Suppose $M = \sup_x l(h(x))$. We define a new loss function $l^M(.) = \frac{1}{M}l(.)$.
$$R_T(h)=\mathbb{E}_{X \sim \mu_{X|T},Y^T \sim \mu_{Y^T|X}} l^{M}(-h(X)Y^T). $$
The corresponding empirical estimator for the expectation above would be 
$$\hat{R}_T(h)=\frac{1}{n_T}\sum_{i \in T}  l^{M}(-h(x_i)Y^T). $$
For the control group, we have 
$$R_C(h)=\mathbb{E}_{X \sim \mu_{X|C},Y^C \sim \mu_{Y^C|X}} l^{M}(-h(X)Y^C). $$
Using estimation of $\mu_{X|T}$ and $\mu_{X|C}$, the corresponding empirical estimator would be:
$$\hat{R}_C(h)=\frac{1}{n_C}\sum_{i \in C} \frac{\mu_{X|T}(x_i)}{\mu_{X|C}(x_i)} l^{M}(-h(x_i)Y^C). $$

\begin{theorem}{
Let $\mathcal{F}$ be a function class, and suppose we have $n$ data points, let $p=Pdim(\mathcal{F})$, the standard pseudo-dimension of $\mathcal{F}$ (see \cite{ARXIV} for precise definition).
Let  $$\Delta_T(\delta)= 2 \sqrt{2\frac{\log S_{\mathcal{F}}(2n_T) + \log \frac{4}{\delta}}{n_T}} \textit{ and}$$ 
 $$\Delta_C(\delta) = 2^{\frac54} \sqrt{d_2(\mu_T||\mu_C)}\sqrt[3/8]{\frac{p\log \frac{2n_Ce}{p} + \log \frac{4}{\delta}}{n_C}}.$$
Then $\forall h \in \mathcal{F},$ with probability at least $1-\delta,$
%\max(R_T(h), R_C(h))
\begin{align*} &\mathbb{E}_{X \sim \mu_X,Y^T \sim \mu_{Y^T|X},Y^C\sim \mu_{Y^C|X}}l_{1}(X,Y^T,Y^C) \\&\leq M \left( \max(\hat{R}_T(h), \hat{R}_C(h))+ \max\left( \Delta_T\left(\frac{\delta}{2}\right), \Delta_C\left(\frac{\delta}{2}\right)\right) \right).\end{align*}
}\end{theorem}
In the theorem, $d_2(P||Q)=2^{D_{KL}(P||Q)}$ where $D_{KL}(P||Q)$ is the usual KL divergence between distributions $P$ and $Q$.

Proof of the generalization bound is provided in the supplementary document. As usual, the bound is algorithm independent.

\section{Experiments}
We cannot observe both treatment and control outcomes for the same observation in real data (this is not standard supervised learning), so ground truth treatment effects must be obtained another way for the purpose of evaluation. 
%\begin{figure}
%   \centering
%   \begin{tabular}{@{}c@{\hspace{.5cm}}c@{}}
%       \includegraphics[page=1,width=1\textwidth]{experimentprotocol.pdf} 
%   \end{tabular}
% \caption{Mechanism for measuring performance of causal inference algorithm }
% \label{fig:Test}
%\end{figure}
%First, let us describe the general experimental setup for the simulated data experiments, which is summarized in Figure \ref{fig:Test}. 
In these experiments, the goal is to test the most basic potential outcomes setting. We randomly assign observations to either the treatment group or control group. We choose distributions $\mu_T$ and $\mu_C$ for generating the $x_i$'s, and choose distributions to generate both potential outcomes $y_i^T$ and $y_i^C$ for each $i$, $y_i^T, y_i^C\in \left\{ -1, 1\right\}$. We  observe either $y_i^T$ or $y_i^C$, depending on whether the observation is in the treatment or control group. The treatment effect $y_i^T-y_i^C$ is thus never observed for any $i$, and takes three possible values: positive, neutral, or negative. We then split the data uniformly into a training set and a test set. The training data were used to build a model that predicts conditional treatment effect given a new test point $x$. We then predict treatment effects for the test data and evaluate our predictions with respect to the ground truth using the conditional difference loss as a performance measure. 

For Causal SVM, we use linear, quadratic, cubic, and radial basis function (RBF) kernels. We compare with matching-based algorithms, such as GenMatch \cite{genmatch} and nearest neighbor matching, followed by ridge regression or kernel ridge regression on the matched groups to create a predictive model. We also compare with algorithms that fit two distinct classification or regression models and take the difference; this includes the difference of ridge regression models, difference of kernel ridge regression models, difference of logistic regression models, difference of SVM models using RBF kernels, and difference of random forests. \textcolor{red}{}Note that the methods where two regression are fitted for different groups are commonly used in meta-algorithms \cite{foster2013subgroup,2017arXiv170603461K}. Also, we compare our algorithms with causal random forests \cite{wager2017estimation}. For methods that involve Genmatch, the pop.size parameter was chosen as $n/2$, and after matching, cross validation was performed for tuning the regularization parameter for the regression methods. For methods involving the difference of two models, cross validation for parameter tuning was performed on the treatment and control data separately.

As discussed earlier, the difference of two distinct classification or regression models is similar to our approach but uses a looser upper bound to the 0-1 loss function: a sum of the terms for treatment and control (as obtained through triangle inequality), rather than a maximum of the two terms. However, using a difference of two models would mean using a strictly richer family of functions to learn the treatment effect. Intuitively one might expect that difference of complex models (e.g. random forests or SVM) would potentially overfit. The bounds are loose enough that it is unclear as to why the sign of the difference of SVM models would necessarily produce useful models, as usually the sign of each single SVM model is used for predicting outcomes.

Our results for each dataset are reported in 2-column tables. The column heading is the value of $\theta$ used in the loss, where $\theta$ is the fraction of data predicted to be neutral. For example when $\theta$=0.1, the 10\% of data with smallest absolute predicted difference are assigned to be neutral. 
 The first $15$ rows of each table are the output of our Causal SVM algorithm. For these methods, the number appended at the end (e.g., 1e-8) indicates the parameter $\gamma$ used. For the RBF kernel, the other number is the inverse kernel width. These are followed by matching based methods, difference of two supervised learning methods, and causal random forests. The two numbers for the causal random forest methods are the $\alpha$ and $\lambda$ parameters in that algorithm. The mean and the standard deviation (in braces) are reported in the table. \textit{The superscript index indicates the rank of the algorithm for the top algorithms.}

%\subsection{\textcolor{blue}{Simulated Dataset}}

\noindent \textbf{Noisy Spirals}

We show the result of a simple but challenging experiment on a causal inference version of the two spiral dataset from \cite{lang1988learning}, which has two covariates.  The goal is to predict a positive treatment effect on one spiral and a negative treatment effect on the other spiral. Half of the training points were randomly assigned to be treated and the other half assigned to control. On one of the two spirals, the treatment effect is positive (treated points had outcome ``yes'' and control points had outcome ``no''), whereas on the other spiral the treatment effect is negative (treated points had outcome ``no'', and control points had outcome ``yes''). We introduce label noise: with probability $20\%$, a data point that should have a positive treatment effect is assigned a negative treatment effect and vice versa. We fit models on the training data, and predicted on out-of-sample test data. Ideally, all points from one spiral should have ``yes'' predictions and the other should have all ``no'' predictions. 

%The \textcolor{blue}{first} data set that we present is the similar to the first spiral data set that we described earlier. The main difference from the first data set is that previously each curve has either positive effect with $100 \%$ certainly or negative effect with $100 \%$ certainty.  We perturn the data set such that on one of the curve, it has positive effect with $80 \%$ certainty and there is another curve with negative effect with $80 \%$  certainty. That is we introduce noise to the ground truth of the outcome. 

\begin{figure*}[htbp]
 \scalebox{0.7}{
   \begin{tabular}{ccccc}
    \centering
    \begin{subfigure}[t]{0.25\textwidth}
        \centering
        \includegraphics[height=1.5in]{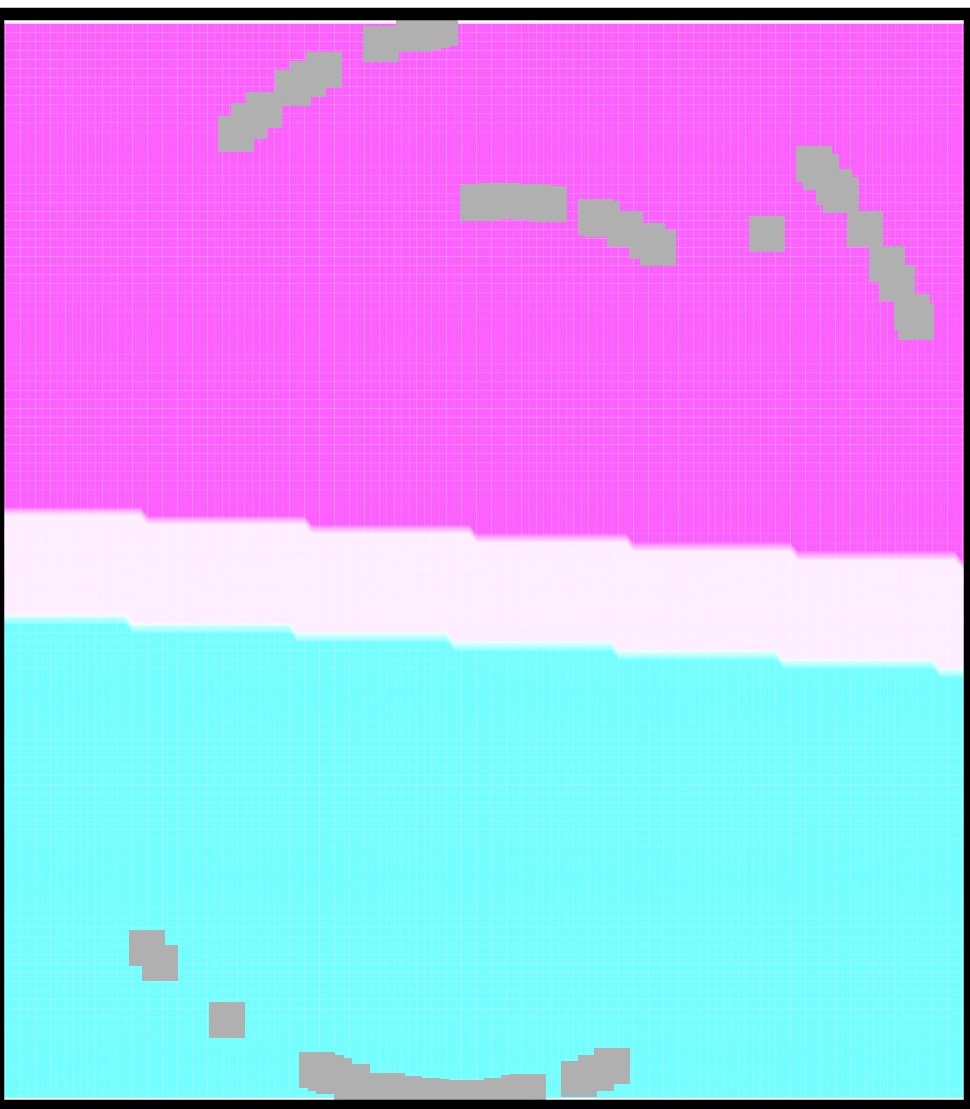}
        \caption{{\large Causal SVM linear kernel}}
    \end{subfigure}%
    &
    \begin{subfigure}[t]{0.25\textwidth}
        \centering
        \includegraphics[height=1.5in]{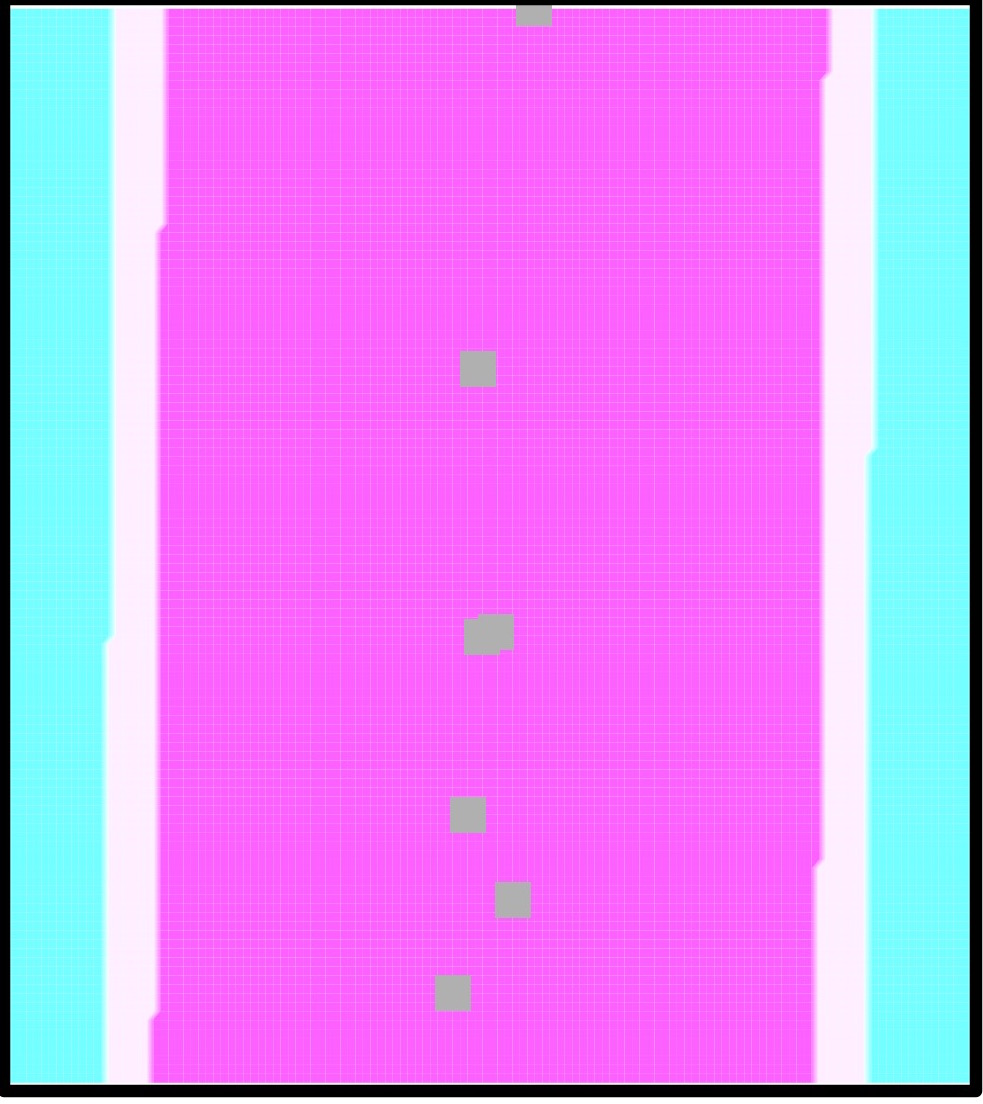}
        \caption{\large Causal SVM quadratic kernel}
    \end{subfigure}
    &
    \begin{subfigure}[t]{0.25\textwidth}
        \centering
        \includegraphics[height=1.5in]{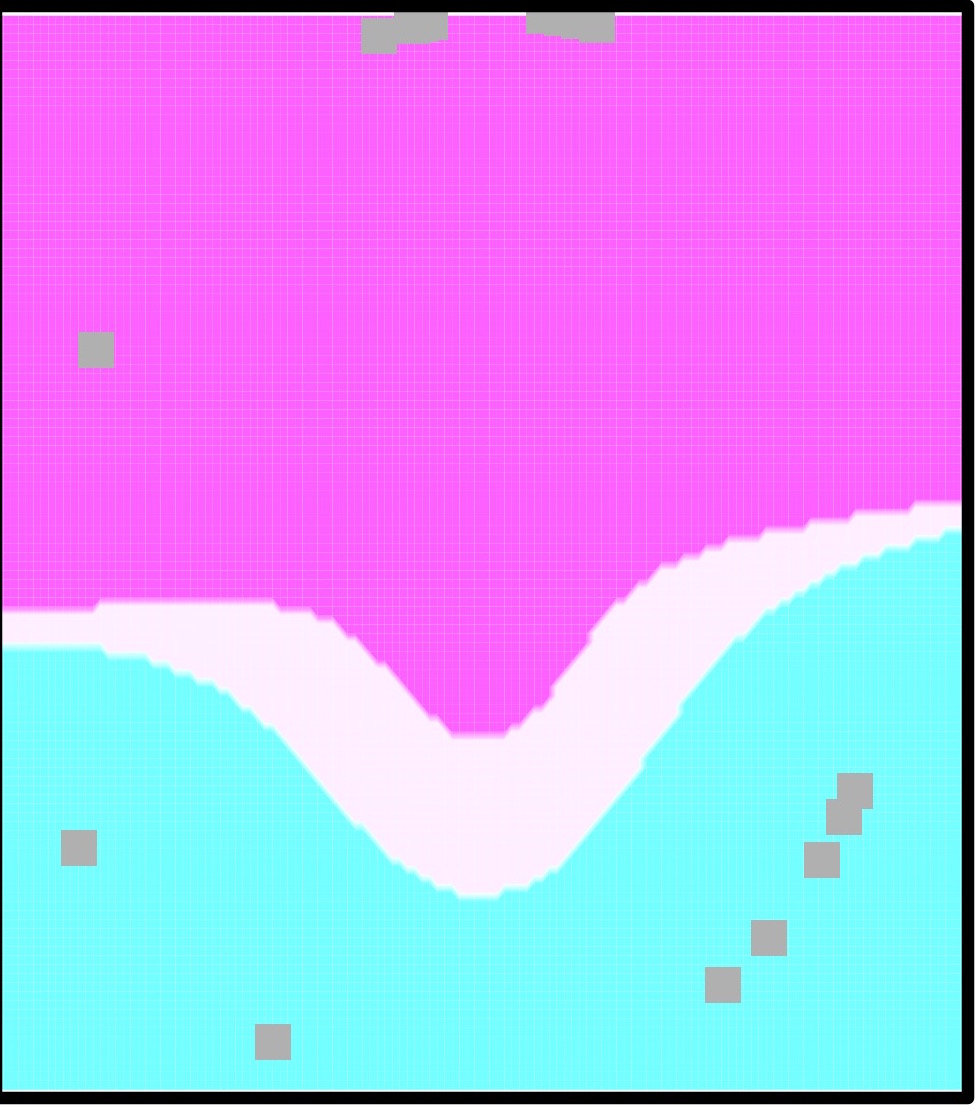}
        \caption{{\large Causal SVM cubic kernel}}
    \end{subfigure}%
    & 
    \begin{subfigure}[t]{0.25\textwidth}
        \centering
        \includegraphics[height=1.5in]{contourrbf2theta1.jpg}
        \caption{{\large Causal SVM RBF kernel}}
    \end{subfigure}
     &
%    \begin{subfigure}[t]{0.25\textwidth}
%        \centering
%        \includegraphics[height=1.5in]{contourgenmatchridgetheta1noise.jpg}
%        \caption{\large Genmatch and ridge regression}
%    \end{subfigure}%
%    \\
%    \begin{subfigure}[t]{0.25\textwidth}
%        \centering
%        \includegraphics[height=1.5in]{contournearestridgetheta1noise.jpg}
%        \caption{\large Nearest neighbor matching and ridge regression}
%    \end{subfigure}
    \begin{subfigure}[t]{0.25\textwidth}
        \centering
        \includegraphics[height=1.5in]{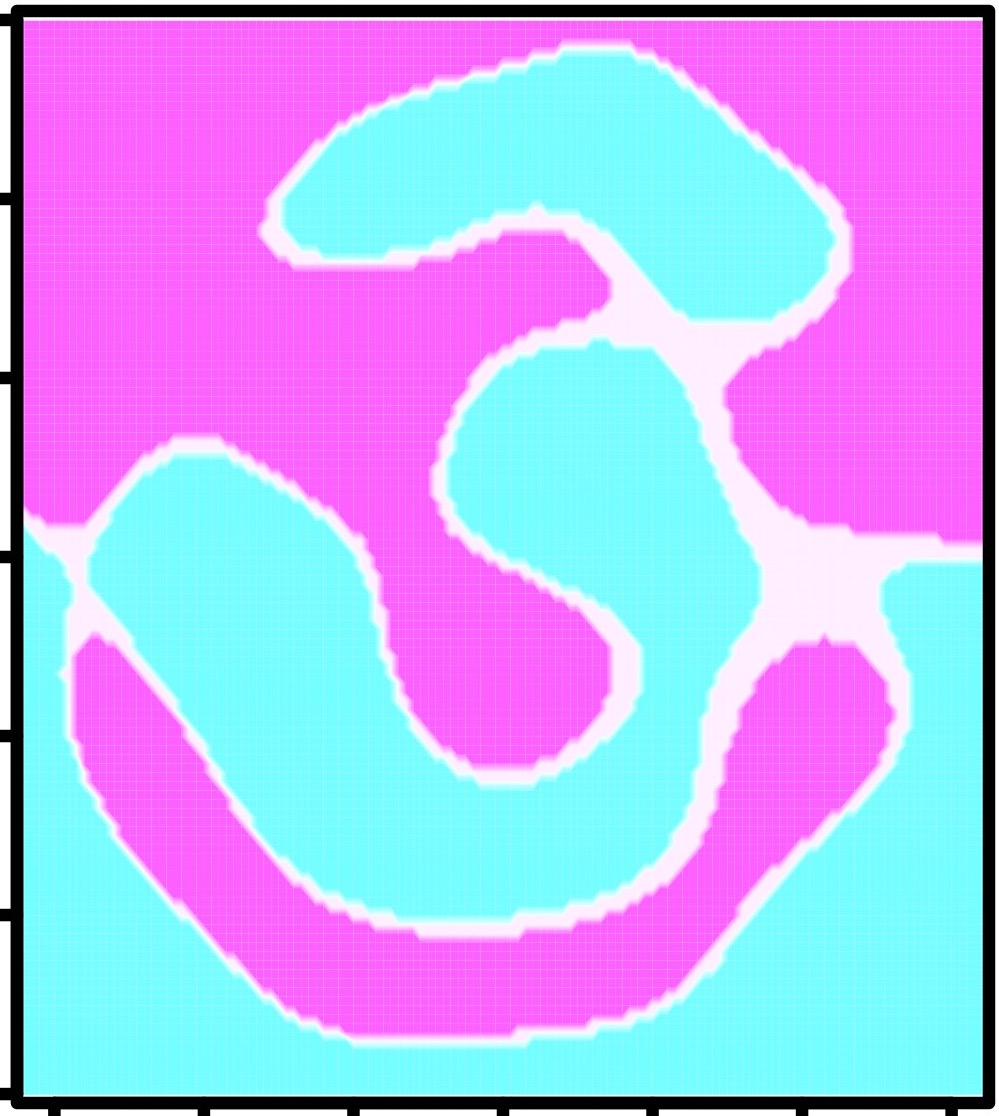}
        \caption{\large Genmatch and kernel ridge regression}
    \end{subfigure}%
    \\
    \begin{subfigure}[t]{0.25\textwidth}
        \centering
        \includegraphics[height=1.5in]{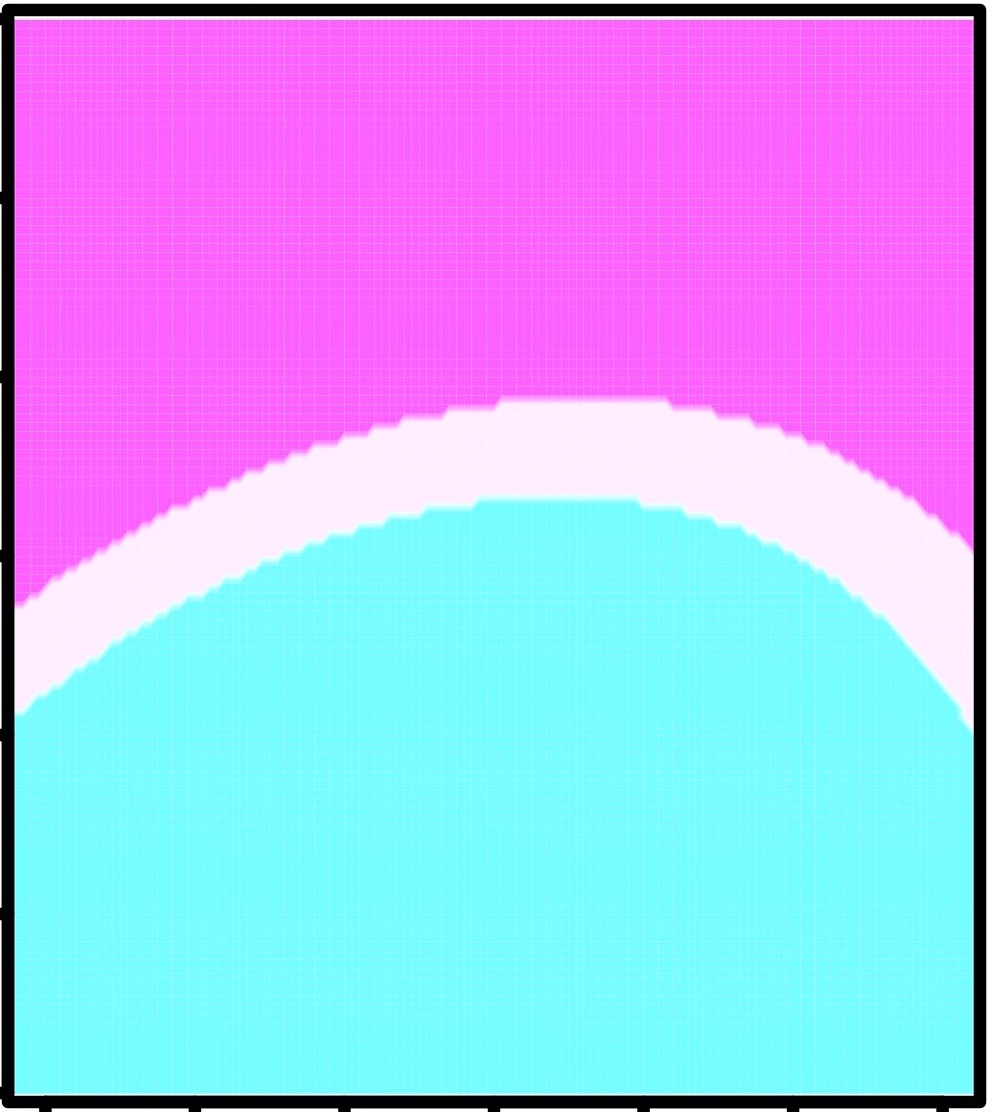}
        \caption{\large Nearest neighbor and kernel ridge regression}
    \end{subfigure}%
    &
%    \begin{subfigure}[t]{0.25\textwidth}
%        \centering
%        \includegraphics[height=1.5in]{contourtworidgetheta1noise.jpg}
%        \caption{Difference of $2$ ridge regression}
%    \end{subfigure}
%    &
    \begin{subfigure}[t]{0.25\textwidth}
        \centering
        \includegraphics[height=1.5in]{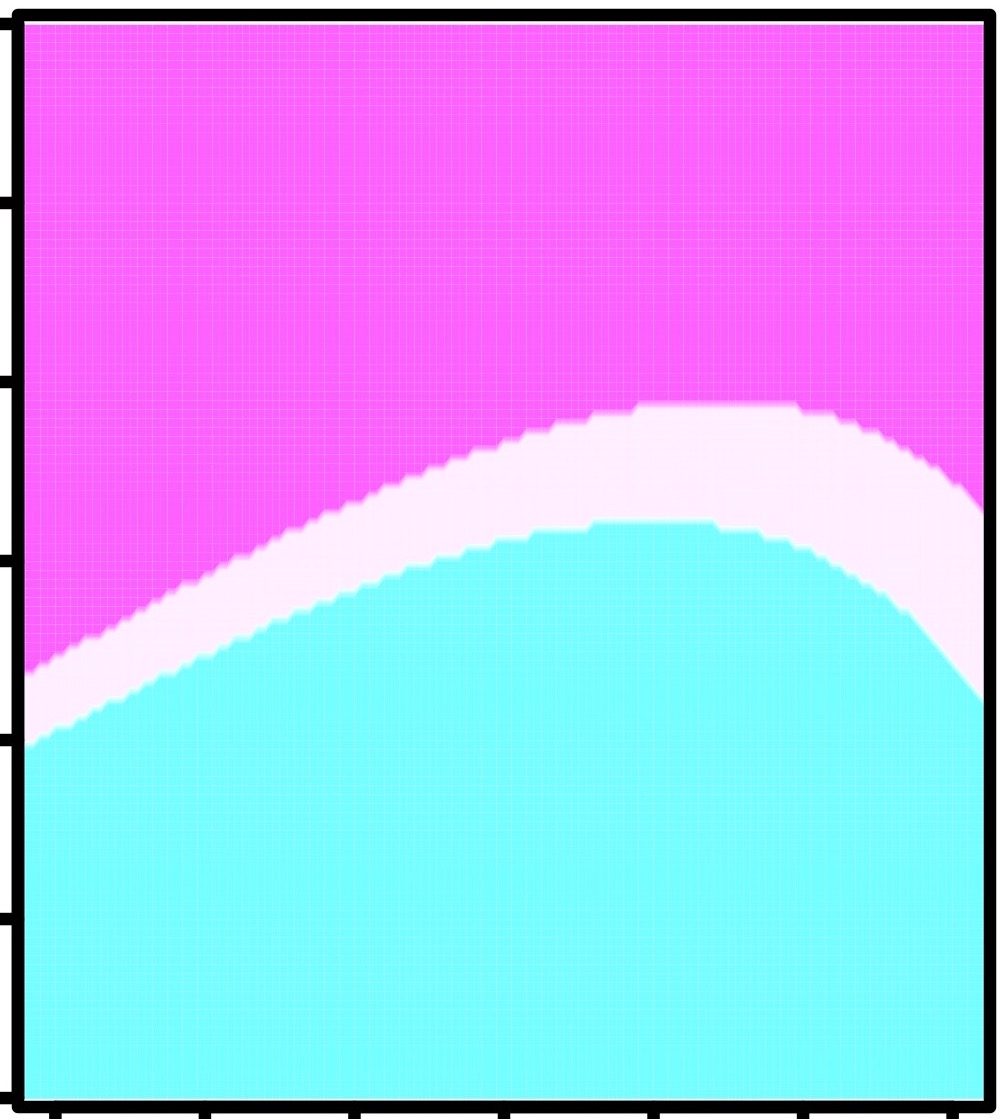}
        \caption{\large Difference of $2$ kernel ridge regression}
    \end{subfigure}
    &
    \begin{subfigure}[t]{0.25\textwidth}
        \centering
        \includegraphics[height=1.5in]{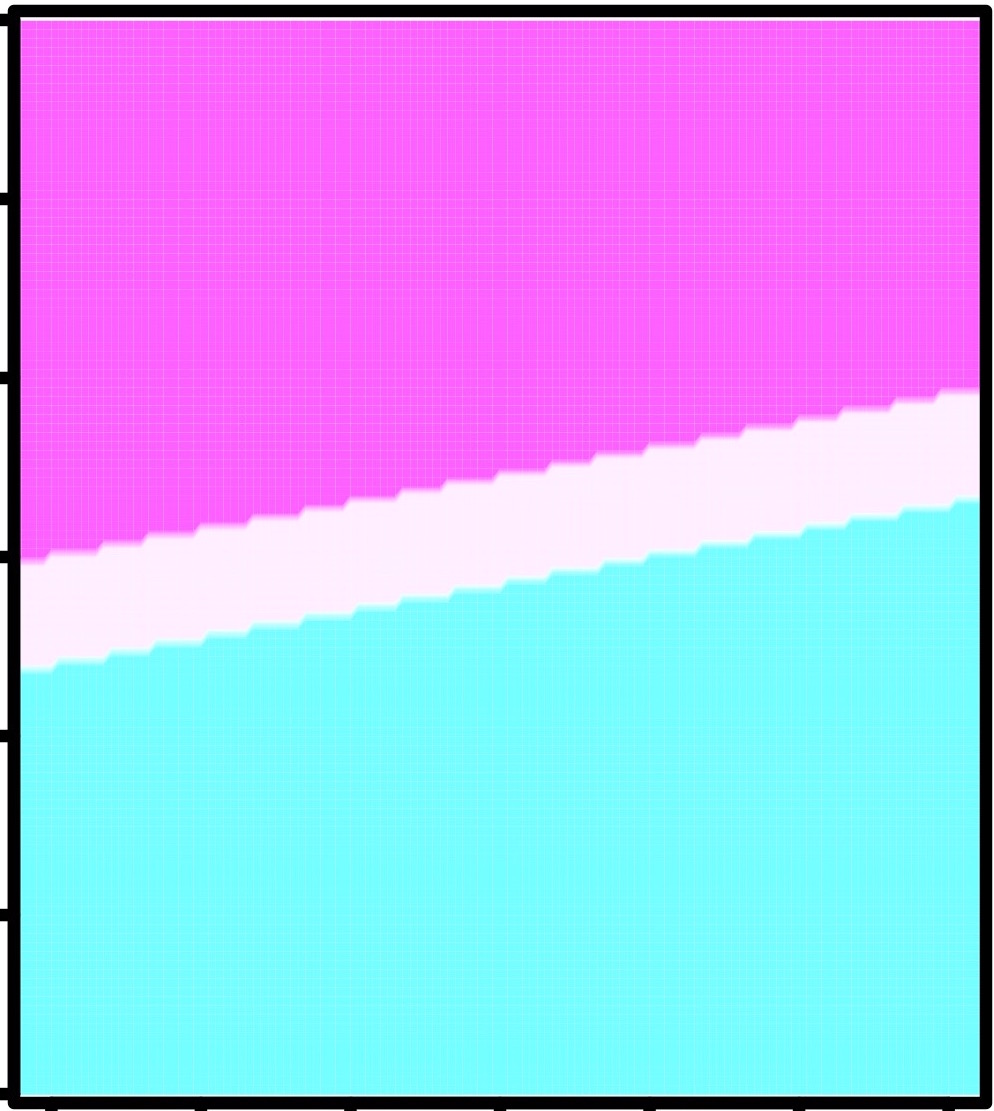}
        \caption{\large Difference of two logistic regression}
    \end{subfigure}%
    & 
    \begin{subfigure}[t]{0.25\textwidth}
        \centering
        \includegraphics[height=1.5in]{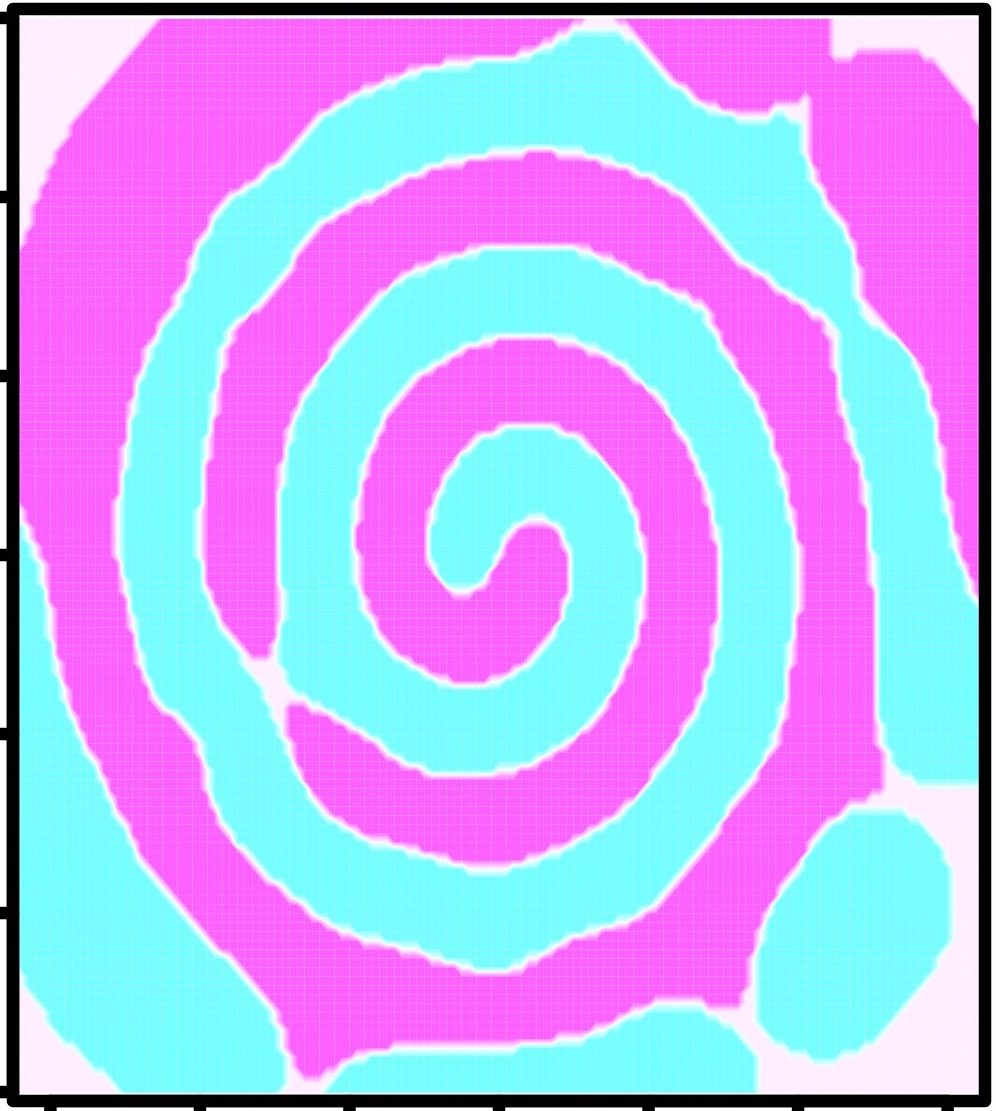}
        \caption{\large Difference of $2$ SVM with RBF}
    \end{subfigure}
    &
    \begin{subfigure}[t]{0.25\textwidth}
        \centering
        \includegraphics[height=1.5in]{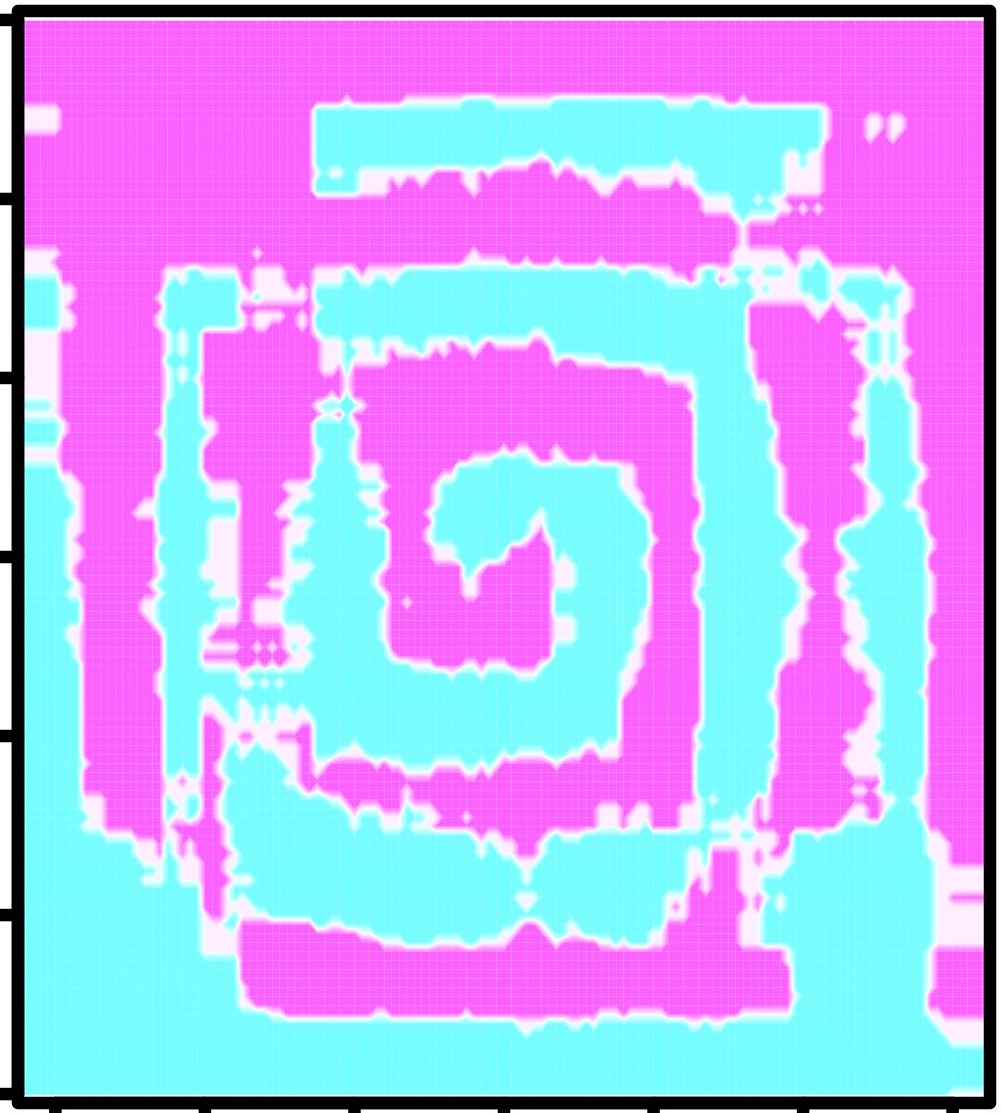}
        \caption{\large Difference of two random forest model}
    \end{subfigure}%
    \end{tabular}}
    \caption{Contour plots showing predicted treatment effect for spiral data with $20\%$ label noise. Support vectors are noted with gray dots for the SVM models. \label{contourspiraltheta1}}
\end{figure*}

\begin{table}[ht]
\centering
\small
\begin{tabular}{rll}
  \hline
$\theta$ & 0.01 & 0.1 \\ 
  \hline
linear Causal SVM 1e-8 & 57.91(1.94) & 52.21(1.91)  \\ 
  linear Causal SVM 1e-6 & 57.91(1.94) & 52.21(1.91) \\ 
  linear Causal SVM 1e-4 & 57.91(1.94) & 52.21(1.91)  \\ 
  quadratic Causal SVM 1e-8 & 58.77(0.95) & 53.51(0.86)  \\ 
  quadratic Causal SVM 1e-6 & 58.77(0.95) & 53.51(0.86)  \\ 
  quadratic Causal SVM 1e-4 & 58.81(1) & 53.52(0.87)  \\ 
  cubic Causal SVM 1e-8 & 56.02(1.72) & 50.51(1.28) \\ 
  cubic Causal SVM 1e-6 & 56.02(1.72) & 50.51(1.28) \\ 
  cubic Causal SVM 1e-4 & 56.23(1.81) & 50.42(1.25) \\ 
  RBF Causal SVM 0.05,  1e-8 & 41.4(2.63) & 36.53(2.46)  \\ 
  RBF Causal SVM 0.05, 1e-6 & 45.52(2.02) & 40.65(1.75)  \\ 
  RBF Causal SVM 0.05, 1e-4 & 55.57(1.7) & 50.28(1.17)  \\ 
  RBF Causal SVM 0.1, 1e-8 & $23.57(0.99)^2$ & $19.45(0.88)^2$  \\ 
  RBF Causal SVM 0.1, 1e-6 & 41.81(2.15) & 37.03(2.13)  \\ 
  RBF Causal SVM 0.1, 1e-4 & 52.47(1.2) & 47.16(1.24)  \\ 
  GenMatch, Ridge & 57.44(1.11) & 52.31(1.1)  \\ 
  Nearest, Ridge & 56.94(1.57) & 51.8(0.93) \\ 
  Genmatch, kernel ridge & 51.7(3.41) & 46.9(3.28)  \\ 
  Nearest, kernel ridge & 52.64(3.2) & 47.27(2.96)  \\ 
  2 ridge & 58.12(1.25) & 52.52(1.09)  \\ 
  2 kernel ridge & 53.04(3.03) & 48.03(2.65) \\ 
  2 logistic & 58.26(0.98) & 52.72(0.92)  \\ 
  2 SVM & $19.91(0.85)^1$ & $17.17(1.17)^1$  \\ 
  2 RF & $25.61(1.32)^3$ & $21.29(1.11)^3$  \\ 
  causal\_rf 0.05 0 & 46.61(1.8) & 41.5(1.6) \\ 
  causal\_rf 0.01 0 & 55.37(1.65) & 50.14(1.86)  \\ 
  causal\_rf 0.05 0.1 & 46.67(1.54) & 41.64(1.79)  \\ 
  causal\_rf 0.01 0.1 & 54.63(2.22) & 49.43(2.05)  \\ 
   \hline
\end{tabular}\caption{Loss values $l_{.01}$ and $l_{.1}$ for spiral data with noise. The best three performers in each column are indicated with superscripts 1, 2 and 3.}\label{SpiralNoise}
\end{table}

As we can see from Table \ref{SpiralNoise}, the linear, quadratic, and cubic methods all perform poorly, because spirals cannot be modeled accurately using linear models or low dimensional polynomials; this would have been clear before performing the experiment, but provides useful baselines. The best performers are RBF SVM models and difference of 2 random forests. The matching methods GenMatch and nearest neighbor seem to have consistently poor performance, as does causal random forests. In fact the performance of these methods is as bad as results obtained from modeling the spirals with linear models.
Figure \ref{contourspiraltheta1} shows models from several machine learning methods. 

More experimental results are in the supplementary materials.

\section{Breaking the Cycle of Drugs and Crime}

Next, we apply our method to data from a social program in the United States, known as Breaking the Cycle (BTC)\cite{BTC}, which studies the effect of intervention on the reduction of crime and drug use. These data were chosen for their relevance to treatment programs for the current opioid epidemic in the U.S. As far as we know, these data have not been previously studied using machine learning techniques. We focus on estimating the effect of the program on reducing non-drug-related crime in Birmingham, Alabama, between years $1997$ and $2001$, based on quality of data and experimental design.  The BTC strategy was to screen offenders shortly after arrest and require those found to use drugs to participate in a drug intervention while under criminal justice supervision. The control group consisted of similar defendants arrested in the year prior to the implementation of BTC. BTC targeted all adult felony defendants and was not limited to those charged with drug offenses. Defendants were ordered to report to BTC for drug screening as a condition of pretrial release. Those who reported drug use, tested positive for drugs, or were arrested on drug felony charges were placed in drug testing and, when appropriate, referred to drug treatment or drug education classes. 

We chose categorical features whose data seemed reliable, that had no missing values, and that had a correlation with the outcome of non-drug-related crime of at least 0.1. We did not use data recorded during the time period over which the outcome was generated, as we intended to build a prediction model for the outcome during that same time period. 
%For example, if someone is arrested when we survey whether someone is correlated, such features are dropped as such feature is not predictive. 
%We only use features that have their correlation with the non-drug related crime of at least absolute value $0.1$. 
The features include: whether the defendant has a drivers license, whether the defendant has access to a automobile, whether an SSI benefit is being received, whether the defendant lives with anyone with an alcohol problem or takes nonprescription drugs, whether the defendant has problems getting along with their father, whether they have suffered for depression within the past $30$ days, whether they have had depression or anxiety for a long period of time, and whether they have trouble understanding. In this dataset, some participants were subsequently dropped from the study as they were later determined to be ineligible, leaving us with $382$ participants.

Our algorithm requires a choice of regularization parameter and kernel parameter. In regular supervised learning, nested cross validation would be the natural method to tune parameters. Typically for causal inference applications, since ground truth is not known, parameters cannot be tuned. In the case of our method, the objective function does \textit{not} require the ground truth to be known, hence, we \textit{can} perform nested cross validation to select parameters using our objective function. 

After running our algorithm,
we wish to examine the results by determining which subgroups benefit from the BTC program. Grouping together the observations with neural and negative estimated treatment effect, to distinguish them from the observations for which the treatment was effective, we used interpretable modeling methods to understand the result. 
% We used  apriori rules and we can also use CORELS (\textcolor{red}{citation}) to build a rule list (a type of decision tree) to make our result more interpretable.

We generated association rules using the Apriori algorithm \cite{JSSv014i15}. The rules indicate that those with access to an automobile seem to benefit from the BTC program in terms of reduction of non-drug related crime. (Having a license could be an indicator of competence of several forms.) We list some of the rules in Table \ref{apriori}.

{\footnotesize

\begin{center}
\begin{table}[!htbp]
\begin{tabular}{ | p{6cm}| c| c | c|c|}  \hline
Antecedent & Effective? & Support & Confidence & Lift \\ \hline
have-automobile=1,                                                                     
       prob-getting-along-father=0,                                                           
       serious-depression-30-days=0 & Y & $0.2173$ &  $1.0000$ & $2.2209$\\ \hline
have-license=1,                                                                        
       serious-depression-30-days=0,                                                          
       serious-depression-life=0 & Y & $0.2539$ &  $0.9899$ & $2.1983$    \\ \hline
have-license=1,                                                                        
       prob-getting-along-father=0,                                                           
       serious-depression-30-days=0 & Y &       $0.2539$ &  $0.9898$ &  $2.1983$    \\ \hline
have-automobile=1,                                                                     
       serious-depression-30-days=0,                                                          
       serious-depression-life=0 &  Y & $0.2173$ &  $0.9881$ & $2.1945$ \\ \hline
have-license=0,                                                                        
       SSI-benefit=0,                                                                         
       prob-getting-along-father=0 & N &  $0.4424$ &   $0.9037$ & $1.6440$   \\ \hline
have-license=0,                                                                        
       prob-getting-along-father=0,                                                           
       trouble-understanding-life=0  & N & $0.3953$  & $0.8935$  & $1.6253$ \\ \hline
have-license=0,                                                                        
       prob-getting-along-father=0,                                                           
       serious-depression-life=1 &  N & $0.1414$ &  $0.8852$ &  $1.6103$ \\ \hline
have-license=0,                                                                        
       live-w-anyone-alcohol=0,                                                               
       prob-getting-along-father=0 &    N &  $0.4398$ & $0.8660$ &  $1.5753$   \\ \hline
have-license=0,                                                                        
       SSI-benefit=0,                                                                         
       trouble-understanding-life=0 &     N & $0.4031$ &  $0.8652$ & $1.5738$   \\ \hline 
\end{tabular} \caption{Association rules for the estimated (Causal SVM) treatment effect of the BTC program on the reduction of non-drug related offenses.}
\label{apriori}
\end{table}  
\end{center}
}

We then created a one-sided decision tree (decision list, or rule list) as an interpretable approximation to the Causal SVM output. We used the CORELS algorithm to produce this rule list \cite{Angelino:2017:LCO:3097983.3098047}, which also certified that this model is optimal according to the objective of accuracy and sparsity on the training set. The rule list is below.

{\footnotesize
\begin{verbatim} 
if (have_drivers_license) then (effective) (87%/13%)
else if (long_term_serious_depression) then (not_effective) (20%/80%)
else if (long_term_trouble_understanding) then (effective) (92%/8%)
else if (SSI_benefit) then (effective) (100%/0%)
else if (prob_getting_along_with_father) then (effective) (91%/9%)
else (not_effective) (3%/97%)\end{verbatim}}

Let us explain the expressions such as $(87\%/13\%)$ on the right of each rule: in the first rule, $87\%$ of observations captured by this rule have an estimated positive treatment effect from Causal SVM. $13\%$ is the percentage of units captured with a negative or neutral estimated treatment effect. 

\section{Discussion}

We presented a framework for estimating personalized treatment effects with theoretically appealing properties. Its surrogate loss bound is tighter than the sum of losses for treatment and control groups. Since it uses global convex optimization, it is easier to troubleshoot and tune than methods that involve greedy splitting, pruning, and averaging (e.g., random forests). Its high-quality experimental results seem to be robust to different datasets, unlike several other methods, meaning that it might be more more trustworthy across domains. Our experiments indicate that it could be useful to include the Causal SVM algorithm in experimental studies, in addition to the algorithms based on separate treatment and control models. The principles used to derive the Causal SVM framework are its surrogate loss definition and bounds, which are of independent interest for other causal inference problems. The generalization bounds are algorithm-independent, and can be applied to any surrogate for the minimax conditional difference loss introduced in this work.  
%We have proposed a family of upper bound for the $0-1$ causal inference loss function. The upper bound is a quantity that can be computable as it doesn't involved manipulation of treatment and control group data simultaneously. This avoided the problem of matching treatment and control data in high dimensional space which is known to be a challenging task. 
%By choosing the hinge loss to create an upper bound, we then derive a quadratic programming problem that is SVM like. Simulation result shows that it is an algorithm that is comparable to existing algorithms. 

%\subsubsection*{Acknowledgements}
\noindent \textbf{Code:} $<$
\url{https://github.com/shangtai/githubcausalsvm}$>$.
\newpage

\bibliographystyle{sapthesis}
\bibliography{AISTATbib}
\newpage
\appendix
\setcounter{secnumdepth}{0}
\section{Appendix}
%\begin{center}\Large
%Causal SVM: Appendix
%\end{center}

%\title{Causal SVM: \\Supplement}

%\author{\name Siong Thye Goh \email stgoh@mit.edu \\
%       \addr Massachusetts Institute of Technology\\
%       %University of Washington\\
%        Cambridge MA ??, USA
%       \AND
%       \name Cynthia Rudin \email  cynthia@cs.duke.edu \\
%       \addr Duke University \\ Durham, NC 27708, USA.}

%\editor{}

%\maketitle

%\newpage
%\section{Experimental  Setting}

%\begin{figure}
%   \centering
%   \begin{tabular}{@{}c@{\hspace{.1cm}}c@{}}
%       \includegraphics[page=1,width=1\textwidth]{experimentprotocol.pdf} 
%   \end{tabular}
% \caption{Mechanism for measuring performance of causal inference algorithm }
% \label{fig:Test}
%\end{figure}
% \newpage

\section{Proof of Theorem 1}
%%%%%%%%%%%%%%%%%%%%%%%%%
\begin{proof}(Of Theorem $1$).
We break down the proofs into five steps for readability.

\subsection{1. Obtaining lower bounds for $\mathbb{E}_{X \sim \mu_{X|T},Y^T \sim \mu_{Y^T|X}}  l(-h(X)Y^T)$ and  $\mathbb{E}_{X \sim \mu_{X|C}, Y^C \sim \mu_{Y^C|X}} \frac{ l(h(X)Y^C) }{\mu_{X|C}(X)/\mu_{X|T}(X)}$.}

\begin{align*}
%\mathbb{E}_{X \sim \mu_T,Y^T \sim \mu_{Y^T|X}} l(-h(X)Y^T)  \\
&\mathbb{E}_{X \sim \mu_{X|T},Y^T \sim \mu_{Y^T|X}, Y^C \sim \mu_{Y^C|X}}  l(-h(X)Y^T)  \\
&=  \mathbb{E}_{X \sim \mu_{X|T},Y^T \sim \mu_{Y^T|X}, Y^C \sim \mu_{Y^C|X}:Y^T>Y^C} l(-h(X)Y^T)  \\&+ \mathbb{E}_{X \sim \mu_{X|T},Y^T \sim \mu_{Y^T|X}, Y^C \sim \mu_{Y^C|X}:Y^T<Y^C}  l(-h(X)Y^T)   \\&+ \mathbb{E}_{X \sim \mu_{X|T},Y^T \sim \mu_{Y^T|X}, Y^C \sim \mu_{Y^C|X}:Y^T=Y^C} l(-h(X)Y^T)   \\
&=  \mathbb{E}_{X \sim \mu_{X|T},Y^T \sim \mu_{Y^T|X}, Y^C \sim \mu_{Y^C|X}:Y^T>Y^C}  l(-h(X))   + \mathbb{E}_{X \sim \mu_{X|T},Y^T \sim \mu_{Y^T|X}, Y^C \sim \mu_{Y^C|X}:Y^T<Y^C}  l(h(X)) \\&+ \mathbb{E}_{X \sim \mu_{X|T},Y^T \sim \mu_{Y^T|X}, Y^C \sim \mu_{Y^C|X}:Y^T=Y^C}  l(-h(X)Y^T)   \\
&\geq \mathbb{E}_{X \sim \mu_{X|T},Y^T \sim \mu_{Y^T|X}, Y^C \sim \mu_{Y^C|X}:Y^T>Y^C} \mathbbm{1}_{h(X) \leq 0}  + \mathbb{E}_{X \sim \mu_{X|T},Y^T \sim \mu_{Y^T|X}, Y^C \sim \mu_{Y^C|X}:Y^T<Y^C}\mathbbm{1}_{h(X) \geq 0} \\&+ \mathbb{E}_{X \sim \mu_{X|T},Y^T \sim \mu_{Y^T|X}, Y^C \sim \mu_{Y^C|X}:Y^T=Y^C}%\lfloor 
l(-h(X)Y^T) 
%\rfloor_+ 
,
\end{align*}
where for the second equation, the reasoning is if $Y^T>Y^C$ then $Y^T=1$, and if $Y^T<Y^C$ then $Y^T=-1.$
The last inequality makes use of the property that if $h(X) \leq 0$, then $l(-h(X)) \geq 1.$ Similarly, if $h(X) \geq 0$, then $l(h(X)) \geq 1.$ 
%Also the loss $l$ is non-negative since it is larger than an indicator function.

By similar reasoning, we have

\begin{align*}
&\mathbb{E}_{X \sim \mu_{X|C},Y^C \sim \mu_{Y^C|X}} \frac{
%\lfloor 
l(h(X)Y^C) 
%\rfloor_+
}{\mu_{X|C}(X)/\mu_{X|T}(X)} \\
&= \mathbb{E}_{X \sim \mu_{X|T},Y^T \sim \mu_{Y^T|X}, Y^C \sim \mu_{Y^C|X}} l(h(X)Y^C)  \\
&=  \mathbb{E}_{X \sim \mu_{X|T},Y^T \sim \mu_{Y^T|X}, Y^C \sim \mu_{Y^C|X}:Y^T>Y^C} l(h(X)Y^C)    + \mathbb{E}_{X \sim \mu_{X|T},Y^T \sim \mu_{Y^T|X}, Y^C \sim \mu_{Y^C|X}:Y^T<Y^C} l(h(X)Y^C)   \\&+ \mathbb{E}_{X \sim \mu_{X|T},Y^T \sim \mu_{Y^T|X}, Y^C \sim \mu_{Y^C|X}:Y^T=Y^C} l(h(X)Y^C)   \\
&=  \mathbb{E}_{X \sim \mu_{X|T},Y^T \sim \mu_{Y^T|X}, Y^C \sim \mu_{Y^C|X}:Y^T>Y^C}  l(-h(X))   + \mathbb{E}_{X \sim \mu_{X|T},Y^T \sim \mu_{Y^T|X}, Y^C \sim \mu_{Y^C|X}:Y^T<Y^C}  l(h(X) ) \\& + \mathbb{E}_{X \sim \mu_{X|T},Y^T \sim \mu_{Y^T|X}, Y^C \sim \mu_{Y^C|X}:Y^T=Y^C}  l(h(X)Y^C)  \\
&\geq \mathbb{E}_{X \sim \mu_{X|T},Y^T \sim \mu_{Y^T|X}, Y^C \sim \mu_{Y^C|X}:Y^T>Y^C} \mathbbm{1}_{h(X)\leq 0}  + \mathbb{E}_{X \sim \mu_{X|T},Y^T \sim \mu_{Y^T|X}, Y^C \sim \mu_{Y^C|X}:Y^T<Y^C} \mathbbm{1}_{h(X) \geq 0} \\&+ \mathbb{E}_{X \sim \mu_{X|T},Y^T \sim \mu_{Y^T|X}, Y^C \sim \mu_{Y^C|X}:Y^T=Y^C} l(h(X)Y^C). 
\end{align*}

\subsection{2. Finding a lower bound for $\max \left( \mathbb{E}_{X \sim \mu_{X|T},Y^T \sim \mu_{Y^T|X}} l(-h(X)Y^T),  \right. \\ \left. \mathbb{E}_{X \sim \mu_{X|C}, Y^C \sim \mu_{Y^C|X}} \frac{ l(h(X)Y^C)}{\mu_C(X)/\mu_T(X)}\right)  $}

We use the property that $a \geq b$ and $c \geq d$ imply $\max(a,c) \geq \max(b,d)$. Taking the maximum of the previous two inequalities in the previous part of the proof, we have
\begin{align*}
&\max \left( \mathbb{E}_{X \sim \mu_{X|T},Y^T \sim \mu_{Y^T|X}}  l(-h(X)Y^T) ,   \mathbb{E}_{X \sim \mu_{X|C}, Y^C \sim \mu_{Y^C|X}} \frac{ l(h(X)Y^C) }{\mu_{X|C}(X)/\mu_{X|T}(X)}\right) \\
& \geq \max \left(  \mathbb{E}_{X \sim \mu_{X|T},Y^T \sim \mu_{Y^T|X}, Y^C \sim \mu_{Y^C|X}:Y^T>Y^C} \mathbbm{1}_{h(X)\leq 0}  + \mathbb{E}_{X \sim \mu_{X|T},Y^T \sim \mu_{Y^T|X}, Y^C \sim \mu_{Y^C|X}:Y^T<Y^C}\mathbbm{1}_{h(X)\geq 0}   \right. \\& \left.+ \mathbb{E}_{X \sim \mu_{X|T},Y^T \sim \mu_{Y^T|X}, Y^C \sim \mu_{Y^C|X}:Y^T=Y^C} l(-h(X)Y^T)  ,   \mathbb{E}_{X \sim \mu_{X|T},Y^T \sim \mu_{Y^T|X}, Y^C \sim \mu_{Y^C|X}:Y^T>Y^C} \mathbbm{1}_{h(X)\leq 0}  \right. \\& +  \mathbb{E}_{X \sim \mu_{X|T},Y^T \sim \mu_{Y^T|X}, Y^C \sim \mu_{Y^C|X}:Y^T<Y^C} \mathbbm{1}_{h(X)\geq 0}  \left.+ \mathbb{E}_{X \sim \mu_{X|T},Y^T \sim \mu_{Y^T|X}, Y^C \sim \mu_{Y^C|X}:Y^T=Y^C} l(h(X)Y^C) \  \right) \\
& \geq \mathbb{E}_{X \sim \mu_{X|T},Y^T \sim \mu_{Y^T|X}, Y^C \sim \mu_{Y^C|X}:Y^T>Y^C} \mathbbm{1}_{h(X)\leq 0}  + \mathbb{E}_{X \sim \mu_{X|T},Y^T \sim \mu_{Y^T|X}, Y^C \sim \mu_{Y^C|X}:Y^T<Y^C} \mathbbm{1}_{h(X)\geq 0} \\ &+ \max \left( \mathbb{E}_{X \sim \mu_{X|T},Y^T \sim \mu_{Y^T|X}, Y^C \sim \mu_{Y^C|X}:Y^T=Y^C} l(-h(X)Y^T ) ,  \mathbb{E}_{X \sim \mu_{X|T},Y^T \sim \mu_{Y^T|X}, Y^C \sim \mu_{Y^C|X}:Y^T=Y^C}  l(h(X)Y^C) \right),
\end{align*}
where the second inequality is because the first two terms within the maximum are exactly the same.

For the next part, we focus on $$\max \left( \mathbb{E}_{X \sim \mu_{X|T},Y^T \sim \mu_{Y^T|X}, Y^C \sim \mu_{Y^C|X}:Y^T=Y^C} l(-h(X)Y^T)  ,  \mathbb{E}_{X \sim \mu_{X|T},Y^T \sim \mu_{Y^T|X}, Y^C \sim \mu_{Y^C|X}:Y^T=Y^C} l(h(X)Y^C) \right), $$
 that is the case where $Y^T=Y^C.$

\subsection{3. Lower bounds for $\mathbb{E}_{X \sim \mu_{X|T},Y^T \sim \mu_{Y^T|X}, Y^C \sim \mu_{Y^C|X}:Y^T=Y^C} l(-h(X)Y^T)$ and $\mathbb{E}_{X \sim \mu_{X|T},Y^T \sim \mu_{Y^T|X}, Y^C \sim \mu_{Y^C|X}:Y^T=Y^C} l(h(X)Y^C)$.}

Since 
\begin{equation*}
l(-h(X)Y^T)  \geq 0
\end{equation*}
because it is an upper bound for an indicator function, we have the following implications
\begin{eqnarray*}
&Y^T=Y^C=-1 \textrm{ and } h(X) \geq 1 \implies l(-h(X)Y^T)=\left(  l(h(X)) \right) \geq 2 \mathbbm{1}_{\left\{h(x) \geq 1 \right\}}\\
&Y^T=Y^C=1 \textrm{ and } h(X) \leq -1 \implies  l(-h(X)Y^T)=\left( l(-h(X)) \right) \geq 2 \mathbbm{1}_{\left\{h(x) \leq -1\right\}}.
\end{eqnarray*}
We have 
\begin{align*}
%\lefteqn{\mathbb{E}_{X \sim \mu_T,Y^T \sim \mu_{Y^T|X}, Y^C \sim \mu_{Y^C|X}:Y^T=Y^C} l(-h(X)Y^T)}
&\mathbb{E}_{X \sim \mu_{X|T},Y^T \sim \mu_{Y^T|X}, Y^C \sim \mu_{Y^C|X}:Y^T=Y^C} l(-h(X)Y^T) 
\\&\geq   2 \mathbb{P}_{X \sim \mu_{X|},Y^T \sim \mu_{Y^T|X}, Y^C \sim \mu_{Y^C|X}} (Y^T=Y^C=-1, h(X) \geq 1) \\&+ 2 \mathbb{P}_{X \sim \mu_{X|T},Y^T \sim \mu_{Y^T|X}, Y^C \sim \mu_{Y^C|X}} (Y^T=Y^C=1, h(X)\leq -1).
\end{align*}
We have a similar result for the control group,
\begin{align*}
&\mathbb{E}_{X \sim \mu_{X|T},Y^T \sim \mu_{Y^T|X}, Y^C \sim \mu_{Y^C|X}:Y^T=Y^C} l(h(X)Y^C) \\&\geq 2 \mathbb{P}_{X \sim \mu_{X|T},Y^T \sim \mu_{Y^T|X}, Y^C \sim \mu_{Y^C|X}} (Y^T=Y^C=1, h(X) \geq 1) \\&+ 2 \mathbb{P}_{X \sim \mu_{X|T},Y^T \sim \mu_{Y^T|X}, Y^C \sim \mu_{Y^C|X}} (Y^T=Y^C=-1, h(X)\leq -1).
\end{align*}

\subsection{4. Lower Bound for the maximum between \\$\mathbb{E}_{X \sim \mu_{X|T},Y^T \sim \mu_{Y^T|X}, Y^C \sim \mu_{Y^C|X}:Y^T=Y^C} l(-h(X)Y^T) $ and  \\$\mathbb{E}_{X \sim \mu_{X|T},Y^T \sim \mu_{Y^T|X}, Y^C \sim \mu_{Y^C|X}:Y^T=Y^C} l(h(X)Y^C) $ }

By using the fact that if $a \geq b$ and $c \geq d$, then we have $\max(a,c) \geq \max(b,d)$ and the result from the previous subsection, we have the first inequality below. The second inequality below is due to  $2max(a,b) \geq a+b.$
\begin{align*}
& \max (\mathbb{E}_{X \sim \mu_{X|T},Y^T \sim \mu_{Y^T|X}, Y^C \sim \mu_{Y^C|X}:Y^T=Y^C} l(-h(X)Y^T)  ,  \\&\mathbb{E}_{X \sim \mu_{X|T},Y^T \sim \mu_{Y^T|X}, Y^C \sim \mu_{Y^C|X}:Y^T=Y^C}  l(h(X)Y^C) )\\
& \geq 2  \max \left( \mathbb{P}_{X \sim \mu_{X|T}, Y^T \sim \mu_{Y^T|X}, Y^C \sim \mu_{Y^C|X}} (Y^T=Y^C=-1, h(X) \leq 1) \right. \\& + \mathbb{P}_{X \sim \mu_{X|T},Y^T \sim \mu_{Y^T|X}, Y^C \sim \mu_{Y^C|X}} (Y^T=Y^C=1, h(X)\leq -1), \\& \mathbb{P}_{X \sim \mu_{X|T},Y^T \sim \mu_{Y^T|X}, Y^C \sim \mu_{Y^C|X}} (Y^T=Y^C=1, h(X) \geq 1) \\&+  \mathbb{P}_{X \sim \mu_{X|T},Y^T \sim \mu_{Y^T|X}, Y^C \sim \mu_{Y^C|X}} (Y^T=Y^C=-1, h(X)\leq -1) \left. \right) \\
& \geq  \mathbb{P}_{X \sim \mu_{X|T},Y^T \sim \mu_{Y^T|X}, Y^C \sim \mu_{Y^C|X}} (Y^T=Y^C=-1, h(X) \geq 1)  \\& + \mathbb{P}_{X \sim \mu_{X|T},Y^T \sim \mu_{Y^T|X}, Y^C \sim \mu_{Y^C|X}} (Y^T=Y^C=1, h(X)\leq -1)  \\  &+\mathbb{P}_{X \sim \mu_{X|T},Y^T \sim \mu_{Y^T|X}, Y^C \sim \mu_{Y^C|X}} (Y^T=Y^C=1, h(X) \geq 1) \\&+  \mathbb{P}_{X \sim \mu_{X|T},Y^T \sim \mu_{Y^T|X}, Y^C \sim \mu_{Y^C|X}} (Y^T=Y^C=-1, h(X)\leq -1)  \\
& = \mathbb{P}_{X \sim \mu_{X|T},Y^T \sim \mu_{Y^T|X}, Y^C \sim \mu_{Y^C|X}} (Y^T=Y^C, |h(X)| \geq 1). 
\end{align*}

\subsection{5. Lower Bound for $\max ( \mathbb{E}_{X \sim \mu_{X|T},Y^T \sim \mu_{Y^T|X}}  l(-h(X)Y^T) ,  \\ \mathbb{E}_{X \sim \mu_{X|C}, Y^C \sim \mu_{Y^C|X}} \frac{ l(h(X)Y^C) }{\mu_C(X)/\mu_T(X)})$}

Combining the result from the seocnd step and fifth step, we have

\begin{align*}
&\max ( \mathbb{E}_{X \sim \mu_{X|T},Y^T \sim \mu_{Y^T|X}} l(-h(X)Y^T) ,  \mathbb{E}_{X \sim \mu_{X|C}, Y^C \sim \mu_{Y^C|X}} \frac{l(h(X)Y^C) }{\mu_{X|C}(X)/\mu_{X|T}(X)} \\
& \geq  \mathbb{E}_{X \sim \mu_{X|T},Y^T \sim \mu_{Y^T|X}, Y^C \sim \mu_{Y^C|X}:Y^T>Y^C} \mathbbm{1}_{h(X)\leq 0}  + \mathbb{E}_{X \sim \mu_{X|T},Y^T \sim \mu_{Y^T|X}, Y^C \sim \mu_{Y^C|X}:Y^T<Y^C} \mathbbm{1}_{h(X)\geq 0} \\ &+ \max \left( \mathbb{E}_{X \sim \mu_{X|T},Y^T \sim \mu_{Y^T|X}, Y^C \sim \mu_{Y^C|X}:Y^T=Y^C} l(-h(X)Y^T)  ,  \mathbb{E}_{X \sim \mu_{X|T},Y^T \sim \mu_{Y^T|X}, Y^C \sim \mu_{Y^C|X}:Y^T=Y^C} l(h(X)Y^C )\right) \\
& \geq \mathbb{E}_{X \sim \mu_{X|T},Y^T \sim \mu_{Y^T|X}, Y^C \sim \mu_{Y^C|X}:Y^T>Y^C} \mathbbm{1}_{h(X)\leq 0}  + \mathbb{E}_{X \sim \mu_{X|T},Y^T \sim \mu_{Y^T|X}, Y^C \sim \mu_{Y^C|X}:Y^T<Y^C} \mathbbm{1}_{h(X)\geq 0}  \\&+ \mathbb{P}_{X \sim \mu_{X|T},Y^T \sim \mu_{Y^T|X}, Y^C \sim \mu_{Y^C|X}} (Y^T=Y^C, |h(X)| \geq 1)  \\
&= \mathbb{E}_{X \sim \mu_{X|T},Y^T \sim \mu_{Y^T|X}, Y^C \sim \mu_{Y^C|X}:Y^T>Y^C} \mathbbm{1}_{h(X)\leq 0}   + \mathbb{E}_{X \sim \mu_{X|T},Y^T \sim \mu_{Y^T|X}, Y^C \sim \mu_{Y^C|X}:Y^T<Y^C} \mathbbm{1}_{h(X)\geq 0}  \\&+ \mathbb{E}_{X \sim \mu_{X|T},Y^T \sim \mu_{Y^T|X}, Y^C \sim \mu_{Y^C|X}: Y^T=Y^C} \mathbbm{1}_{|h(X)| \geq 1} 
\\
& =\mathbb{E}_{X \sim \mu_{X|T},Y^T \sim \mu_{Y^T|X},Y^C\sim \mu_{Y^C|X}}l_{0-1}(X,Y^T,Y^C, h) \\
& \geq \mathbb{E}_{X \sim \mu_{X|T},Y^T \sim \mu_{Y^T|X},Y^C\sim \mu_{Y^C|X}}l_{1}(X,Y^T,Y^C, h).
\end{align*}

That is we have proven that the expectation of the loss function is upper bounded by 

$$\max \left( \mathbb{E}_{X \sim \mu_{X|T},Y^T \sim \mu_{Y^T|X}} l(-h(X)Y^T ), \right. \left. \mathbb{E}_{X \sim \mu_{X|C}, Y^C \sim \mu_{Y^C|X}} \frac{l(h(X)Y^C)}{\mu_{X|C}(X)/\mu_{X|T}(X)}\right). $$
\end{proof}

Remark: The proof of Theorem $2$ is actually included  where we stop just before the final inequality.

\begin{figure*}[htbp]
    \centering
    \begin{subfigure}[t]{0.35\textwidth}
        \centering
        \includegraphics[height=2.4in]{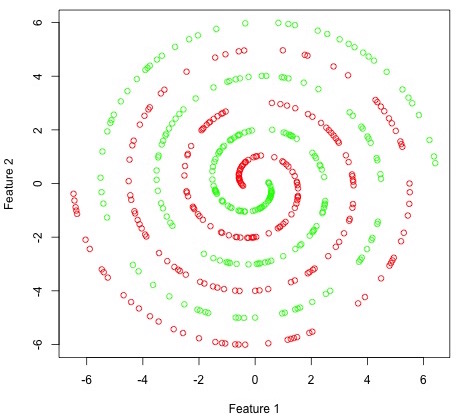}
        \caption{Observed treatment output. The red data points indicate $y^T=1$ while the green data points indicate $y^T=-1$.}
    \end{subfigure}%
    %\newline
    \hspace*{70pt}
    \begin{subfigure}[t]{0.35\textwidth}
        \centering
        \includegraphics[height=2.4in]{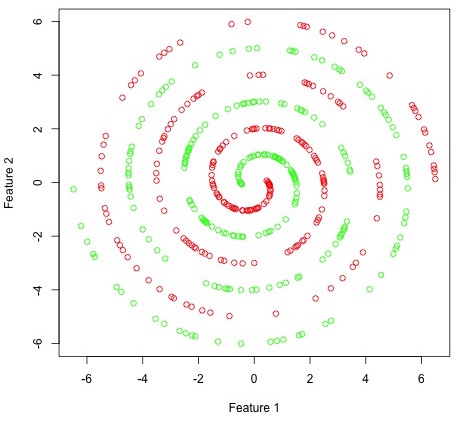}
        \caption{Observed control output. The red data points indicate $y^C=1$ while the green data points indicate $y^C=-1$.}
    \end{subfigure}
    ~
    \begin{subfigure}[t]{0.35\textwidth}
        \centering
        \includegraphics[height=2.4in]{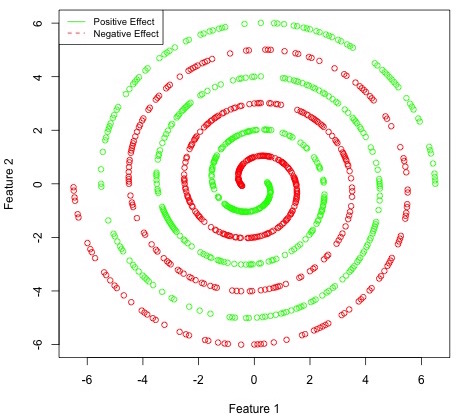}
        \caption{The ground truth treatment effect (not observed). The green data points indicate positive treatment effect and the red data points indicate negative treatment effect.}
    \end{subfigure}
    \caption{The ground truth and the observed outcome for the spiral data set \label{coolspiral}}
\end{figure*}

\section{Proof of Generalization Bound}

From  \cite{bousquet2004introduction}, we have $\forall \delta > 0$, with probability at least $1-\delta$

$$\forall h \in \mathcal{F}, R_T(h) \leq \hat{R}_T(h)+\Delta_T(\delta),$$

where $$\Delta_T(\delta)= 2 \sqrt{2\frac{\log S_{\mathcal{F}}(2n_T) + \log \frac{4}{\delta}}{n_T}}.$$

Unlike conventional statistical learning bounds, recall that we are working with two different distributions, $\mu_T$ and $\mu_C$ of which we have chosen $\mu_T$ to be our target distribution. The use of Radon-Nikodym derivatives to transform $\mu_C$ to $\mu_T$  corresponds to  importance weighting.

%Let $Pdim(U)$ be the pseudo--dimension of a real-valued function class $U$, $L_h(x)=L(h(x),y)$, $w(x)=P(x)/Q(x).$

%$$R(h) = \mathbb{E}_{x \sim P}[L(h(x),y(x)]$$
%and

%$$\hat{R}(h) = \frac{1}{n} \sum_{i=1}^n w(x_i)L(h(x_i),y(x_i))$$
We will build our result on Theorem $3$ in \cite{cortes2010learning} which states the following:

Let $F$ be a hypothesis set such that $Pdim\left(\left\{ L_h(x): h \in \mathcal{F} \right\} \right) = p < \infty$. 
Let $X$ denote the input space and let $Y$ be the label set. We let $L:Y \times Y \to [0,1]$ be a loss function. We let $f:X \to Y$ be the target labeling function. We let $L_h(x)$ denote $L(h(x),f(x))$ in the absence of ambiguity about the target function $f$.

For any hypothesis $h \in \mathcal{F}$, we denote by $R(h)$ its loss and by $\hat{R}_w(h)$ its weighted empirical loss:

\begin{align*}
R(h) &= \mathbb{E}_{x \sim P} [L(h(x)), f(x)] \\
\hat{R}_w(h) &= \frac{1}{m}\sum_{i=1}^m w(x_i) L(h(x_i), f(x_i))
\end{align*}

 Assume that $d_2(P||Q)= 2^{D_{KL}(P||Q)} < + \infty$ and $w(x) \neq 0$ for all $x$. Then, for any $\delta>0$, with probability at least $1-\delta$, the following holds:

\begin{equation}\label{Cortes}
\forall h \in \mathcal{F}, R(h) \leq \hat{R}_w(h) + 2^{\frac54} \sqrt{d_2(P||Q)}\sqrt[3/8]{\frac{p\log \frac{2ne}{p} + \log \frac{4}{\delta}}{n}}.
\end{equation}

%Here $D_2(P||Q)$ refers to the KL divergence between $P$ and $Q$.

From Equation \ref{Cortes}, we can conclude that 

$\forall h \in \mathcal{F}$, with probability at least $1-\delta$, 

 we have $$R_C(h) \leq \hat{R}_C(h)+\Delta_C(\delta),$$

where $$\Delta_C(\delta) = 2^{\frac54} \sqrt{d_2(\mu_T||\mu_C)}\sqrt[3/8]{\frac{p\log \frac{2n_Ce}{p} + \log \frac{4}{\delta}}{n_C}}.$$

Hence, we can combine these two inequalities using union bound and obtain the following:

$\forall h \in \mathcal{F},$ with probability at least $1-\delta,$

\begin{align*}\max(R_T(h), R_C(h)) &\leq \max \left(\hat{R}_T(h)+\Delta_T\left(\frac{\delta}{2}\right), \hat{R}_C(h)+\Delta_C\left(\frac{\delta}{2}\right)\right) \\
&\leq \max(\hat{R}_T(h), \hat{R}_C(h))+ \max\left( \Delta_T\left(\frac{\delta}{2}\right), \Delta_C\left(\frac{\delta}{2}\right)\right)\end{align*}

We complete the proof by noticing that from definition of $l^M$ and linearity of expectation that 
%\ref{theoremupperbd}, 
we have \begin{align*}\mathbb{E}_{X \sim \mu_{X|T},Y^T \sim \mu_{Y^T|X},Y^C\sim \mu_{Y^C|X}}l_{1}(X,Y^T,Y^C, h) &= M \mathbb{E}_{X \sim \mu_{X|T},Y^T \sim \mu_{Y^T|X},Y^C\sim \mu_{Y^C|X}}l_{1}^M(X,Y^T,Y^C, h)\\& \leq M \max(R_T(h), R_C(h))
\\
&\leq \max(\hat{R}_T(h), \hat{R}_C(h))+ \max\left( \Delta_T\left(\frac{\delta}{2}\right), \Delta_C\left(\frac{\delta}{2}\right)\right)
. \end{align*}

\subsection{Additional Experimental Results}

Due to the page limit constraint in the main paper, here are results on some additional data sets.

\subsubsection{Spiral Dataset without noise:}
The first data set that we present is the spiral data set as shown in Figure \ref{coolspiral} without any noise. The support vectors are noted in the figure for causal SVM. Causal SVM and 2 SVM perform comparably, and the determination of which one performs better depends on the kernel bandwidth -- here the 2 SVM is slightly better. Numerical results are in Table \ref{fig:spirallosstheta}.

%The causal SVM  formulation has several clear advantages: because it uses a single model rather than a difference, it predicts the treatment effect directly and thus tends to be more accurate. The model is a solution to a single convex optimization problem, and does not rely on greedy splitting or pruning techniques that can lead to overfitting or other non-robustness issues. Because of the nonlinearities in the spiral data, the best kernel for this application is the RBF kernel. \textcolor{blue}{The best algorithm for this dataset would be the difference of $2$ SVM method. In fact, it almost perfectly solve the problem with a very rich family class. The other method that has similar performance as our method would be the difference of two random forests.Random forests tend to perform well for non-linear problems, but often tend to overfit (though not on these data). Unlike causal SVM, they do not solve a global optimization problem.  }
%The only other model whose accuracy is close to that of causal SVM is the difference of random forests, and both models have nearly perfect accuracy. 

%\textcolor{blue}{Some readings are weird, something is still wrong}

\begin{table}[ht]
\centering \small
\begin{tabular}{rll}
  \hline
 & 0.01 & 0.1 \\ 
  \hline
linear causal SVM 1e-8 & 47.33(1.82) & 43.47(1.29)  \\ 
  linear causal SVM 1e-6 & 47.33(1.82) & 43.47(1.29)  \\ 
  linear causal SVM 1e-4 & 47.34(1.81) & 43.47(1.29)  \\ 
  quadratic causal SVM 1e-8 & 51(0.97) & 46.13(0.81)  \\ 
  quadratic causal SVM 1e-6 & 51(0.97) & 46.13(0.81)  \\ 
  quadratic causal SVM 1e-4 & 50.95(1.01) & 46.16(0.79)  \\ 
  cubic causal SVM 1e-8 & 46.23(2.16) & 42.3(0.82)  \\ 
  cubic causal SVM 1e-6 & 46.23(2.16) & 42.3(0.82)  \\ 
  cubic causal SVM 1e-4 & 46.21(2.34) & 42.31(0.82)  \\ 
  rbf causal SVM 0.05,  1e-8 & 25.27(2.9) & 20.95(2.28)  \\ 
  rbf causal SVM 0.05, 1e-6 & 31.26(2.81) & 26.98(2.98)  \\ 
  rbf causal SVM 0.05, 1e-4 & 47.69(2.42) & 42.94(2.16)  \\ 
  rbf causal SVM 0.1, 1e-8 & $4.32(0.78)^3$ & $2.03(0.27)^2$  \\ 
  rbf causal SVM 0.1, 1e-6 & 26.39(2.63) & 22.3(2.04)  \\ 
  rbf causal SVM 0.1, 1e-4 & 39.25(2.17) & 34.65(1.76) \\ 
  GenMatch, Ridge & 48.34(2.94) & 43.84(1.91)  \\ 
  Nearest, Ridge & 49.33(1.66) & 44.88(1.47) \\ 
  Genmatch, kernel ridge & 43.64(4.94) & 38.69(4.24) \\ 
  Nearest, kernel ridge & 44.93(4.95) & 40.44(5.02)  \\ 
  2 ridge & 48.52(2.33) & 44.27(1.63)  \\ 
  2 kernel ridge & 43.11(3.52) & 38.73(3.22)  \\ 
  2 logistic & 48.74(2.32) & 44.3(1.75) \\ 
  2 SVM & $0.05(0.05)^1$ & $0.01(0.03)^1$ \\ 
  2 RF & $4.24(0.87)^2$ & $1.51(0.76)^2$ \\ 
  causal\_rf 0.05 0 & 34.06(3.32) & 29.72(2.89)  \\ 
  causal\_rf 0.01 0 & 47.04(2.64) & 42.14(2.36)  \\ 
  causal\_rf 0.05 0.1 & 32.92(2.67) & 28.85(2.6) \\ 
  causal\_rf 0.01 0.1 & 45.35(2.15) & 41.37(2.56) \\ 
   \hline
\end{tabular}
\caption{Numerical output for the spiral data set. As we can see, our method is the best method without using difference of two supervised classifiers.}\label{fig:spirallosstheta}
\end{table}

\subsubsection{A Dataset Where the Treatment Effect Changes a Few Times}

We construct a $2$-dimensional data set as follows. The features are distributed uniformly between $0$ and $1$. We denote $x_{i,j}$ as $i$ is the index for the $i$-th data point  and $j$ is the index for the feature. If $x_{i,1} < 0.6$, $y_i^T = 1$ with probability $0.4$ and $y_i^C=1$ with probability $0.6$; If $x_{i,1} $ is between $0.6$ and $0.8,$ $y_i^T = 1$ with probability $0.3$ and $y_i^C=1$ with probability $0.7$;  Otherwise, $y_i^T = 1$ with probability $0.8$ and $y_i^C=1$ with probability $0.2$;

Causal SVM outperforms other algorithms for this data set. Numerical results are in Table \ref{Advantage}.

% latex table generated in R 3.3.2 by xtable 1.8-2 package
% Wed Sep  6 00:44:54 2017

\begin{table}[ht]
\centering\small
\begin{tabular}{rll}
  \hline
 & 0.01 & 0.1 \\ 
  \hline
linear causal SVM 1e-8 & 62.1(3.18) & 55.3(3.3) \\ 
  linear causal SVM 1e-6 & 62.1(3.18) & 55.3(3.3)  \\ 
  linear causal SVM 1e-4 & 61.5(3.34) & 54.6(3.06)  \\ 
  quadratic causal SVM 1e-8 & 61.2(4.54) & 53.9(4.33) \\ 
  quadratic causal SVM 1e-6 & 60.6(4.55) & 53.5(4.33)  \\ 
  quadratic causal SVM 1e-4 & 61.7(5.06) & 55(5.25) \\ 
  cubic causal SVM 1e-8 & $59.6(3.78)^1$ & $53.2(4.02)^2$  \\ 
  cubic causal SVM 1e-6 & 61.8(6) & 55.3(5.29)  \\ 
  cubic causal SVM 1e-4 & 61.5(5.74) & 54.9(5.2)  \\ 
  rbf causal SVM 0.05,  1e-8 & 60.5(3.84) & 53.7(3.59)  \\ 
  rbf causal SVM 0.05, 1e-6 & 60.1(4.65) & $53.4(4.86)^3$  \\ 
  rbf causal SVM 0.05, 1e-4 & 60.7(3.56) & 53.7(3.33)  \\ 
  rbf causal SVM 0.1, 1e-8 & 61.6(3.84) & 54.8(3.88) \\ 
  rbf causal SVM 0.1, 1e-6 & $60(4.22)^2$ & $53.1(4.09)^1$ \\ 
  rbf causal SVM 0.1, 1e-4 & $60.1(3.98)^3$ & 53.6(3.06)  \\ 
  GenMatch, Ridge & 65.2(3.71) & 58.4(3.31)  \\ 
  Nearest, Ridge & 66.2(2.97) & 59.5(2.59)  \\ 
  Genmatch, kernel ridge & 65.6(6.19) & 58.5(5.56)  \\ 
  Nearest, kernel ridge & 65.2(3.52) & 58.9(3.93)  \\ 
  2 ridge & 63(3.5) & 56.4(3.57)  \\ 
  2 kernel ridge & 63.9(3.96) & 57.4(3.81)  \\ 
  2 logistic & 63.3(3.43) & 57.3(2.83)  \\ 
  2 SVM & 66(3.46) & 59.5(3.24) \\ 
  2 RF & 64.5(1.51) & 59(1.89) \\ 
  causal\_rf 0.05 0 & 67.3(6.5) & 59.8(5.81) \\ 
  causal\_rf 0.01 0 & 64.7(23.28) & 63.4(22.81)  \\ 
  causal\_rf 0.05 0.1 & 68.4(6.02) & 61.1(4.61)  \\ 
  causal\_rf 0.01 0.1 & 70.4(4.9) & 70.4(4.9)\\ 
   \hline
\end{tabular} \caption{The output for a data set where the treatment effect changes a few times. Our method seems to be more suited for this type of data set.}\label{Advantage}
\end{table}

% latex table generated in R 3.3.2 by xtable 1.8-2 package

We can see from Table \ref{Advantage} that for this particular data set, our approaches outperform the other algorithms.

 \subsubsection{Imbalanced when it is more likely to belong to the control group}

This is a simulated data set that consists of $1000$ data points. Each data point has $30$ features. The first $20$ features are independently generated from a normal distribution with mean $0$ and variance $1$, while the remaining $10$ features are uniformly distributed between $-1$ and $1$. The treatment effect is determined by the $2$-norm of the feature. If the feature has norm that is bigger than $3$, then $y_i^{T}$ takes value $1$ with probability $0.8$ ; Otherwise, $y_i^T$ takes value $-1$ with probability $0.2$. $y_i^C$ always take value $1$ with probability $0.2$. Each data point has a probability of $0.3$ of being assigned to the control group.
Table \ref{ControlTable} shows that causal SVM is comparable to 2-SVM, and matching-based methods have worse performance.

\begin{table}[ht]
\centering\small
\begin{tabular}{rll}
  \hline
 & 0.01 & 0.1  \\ 
  \hline
linear causal SVM 1e-8 & 57.8(2.92) & 51.6(2.91)  \\ 
  linear causal SVM 1e-6 & 57.8(2.92) & 51.6(2.91)  \\ 
  linear causal SVM 1e-4 & 57.82(2.98) & 51.68(2.91)  \\ 
  quadratic causal SVM 1e-8 & 63.36(1.83) & 57.02(1.61) \\ 
  quadratic causal SVM 1e-6 & 63.4(1.81) & 57.02(1.6) \\ 
  quadratic causal SVM 1e-4 & 59.6(1.35) & 53.3(1.46) \\ 
  cubic causal SVM 1e-8 & 62.58(1.87) & 56.1(1.81) \\  
  cubic causal SVM 1e-6 & 62.58(1.87) & 56.12(1.83) \\ 
  cubic causal SVM 1e-4 & 55.38(2.07) & 49.58(1.9)  \\ 
  rbf causal SVM 0.05,  1e-8 & 55.02(1.6) & 48.82(1.55) \\ 
  rbf causal SVM 0.05, 1e-6 & 55.02(1.6) & 48.82(1.55) \\ 
  rbf causal SVM 0.05, 1e-4 & 55.02(1.6) & 48.82(1.55)  \\ 
  rbf causal SVM 0.1, 1e-8 & 50.82(1.74) & 44.2(1.62) \\ 
  rbf causal SVM 0.1, 1e-6 & 50.82(1.74) & $44.16(1.58)^2$  \\ 
  rbf causal SVM 0.1, 1e-4 & 50.82(1.74) & $44.16(1.58)^2$  \\ 
  GenMatch, Ridge & 52.84(2.56) & 47.22(2.28)  \\ 
  Nearest, Ridge & 50.88(2.23) & 46.38(2.51) \\ 
  Genmatch, kernel ridge & 56.76(2.44) & 50.76(2.2)  \\ 
  Nearest, kernel ridge & 53.52(2.09) & 47.96(1.82)  \\ 
  2 ridge & 51.02(1.97) & 46(2.19)  \\ 
  2 kernel ridge & 53.44(1.94) & 47.7(2.05)  \\ 
  2 logistic & 57.4(1.8) & 50.88(1.53)  \\ 
  2 SVM & $50.4(2.1)^1$ & $43.78(2.07)^1$ \\ 
  2 RF & 50.86(2.12) & 45.84(1.97) \\ 
  causal\_rf 0.05 0 & $50.68(2.02)^3$ & 45.1(2.22)  \\ 
  causal\_rf 0.01 0 & 50.82(2.15) & 45.46(2.04) \\ 
  causal\_rf 0.05 0.1 & $50.66(2.16)^2$ & 45.16(2.04)  \\ 
  causal\_rf 0.01 0.1 & 50.78(2.11) & 45.66(2.5) \\ 
   \hline
\end{tabular} \caption{The output for a data set that simulate a scenario where it is more likely to be assigned to the control group. It is shown that our RBF based methods beat matching based method.} \label{ControlTable}
\end{table}

 \subsubsection{A dataset with treatment effect that changes a few times in high dimensions}

This is a data set which consists of $1000$ data points where each data point consists of $120$ features. $60$ of the features follows independent normal distribution with mena $0$ and standard deviation $1$ and $60$ features follows uniform distribution between $-1$ and $1$. The treatment effect is a function of the $2$-norm of each data point. If the norm is less than $3$, $y_i^T$ takes value $1$ with probability $0.4$ while $y_i^C$ takes value $1$ with probability $0.6$; If the norm is between $3$ and $4$,  $y_i^T$ takes value $1$ with probability $0.3$ while $y_i^C$ takes value $1$ with probability $0.7$; Otherise, $y_i^T$ and $y_i^C$ independently take value $1$ with probability $0.2$. It is equally likely for a data to be assigned to a treatment group or a control group.

\begin{table}[ht]
\centering\small
\begin{tabular}{rll}
  \hline
 & 0.01 & 0.1  \\ 
  \hline
linear causal SVM 1e-8 & 69.52(2.34) & 62.98(2.26)  \\ 
  linear causal SVM 1e-6 & 69.5(2.38) & 62.98(2.26)  \\ 
  linear causal SVM 1e-4 & 69.48(2.2) & 62.9(2.11) \\ 
  quadratic causal SVM 1e-8 & 67.3(1.33) & 60.56(1.32) \\ 
  quadratic causal SVM 1e-6 & 67.2(1.38) & 60.66(1.3)  \\ 
  quadratic causal SVM 1e-4 & 67.18(1.36) & 60.66(1.3) \\ 
  cubic causal SVM 1e-8 & 61.74(1.28) & 55.1(1.44) \\ 
  cubic causal SVM 1e-6 & 61.7(1.32) & $55.08(1.51)^2$\\ 
  cubic causal SVM 1e-4 & 61.7(1.32) & $55.08(1.51)^2$ \\ 
  rbf causal SVM 0.05,  1e-8 & 61.12(1.35) & 55.54(1.05)  \\ 
  rbf causal SVM 0.05, 1e-6 & 61.12(1.35) & 55.54(1.05) \\ 
  rbf causal SVM 0.05, 1e-4 & 61.12(1.35) & 55.54(1.05) \\ 
  rbf causal SVM 0.1, 1e-8 & 61.2(1.38) & 55.82(1.07)  \\ 
  rbf causal SVM 0.1, 1e-6 & 61.2(1.38) & 55.8(1.06) \\ 
  rbf causal SVM 0.1, 1e-4 & 61.2(1.38) & 55.8(1.06)\\ 
  GenMatch, Ridge & 61.74(1.59) & 55.9(1.66)  \\ 
  Nearest, Ridge & 61.6(1.87) & 55.68(1.72)  \\ 
  Genmatch, kernel ridge & 62.98(1.53) & 56.62(1.43) \\ 
  Nearest, kernel ridge & 62.94(1.78) & 56.12(1.67)\\ 
  2 ridge & $61.1(1.54)^3$ & 55.66(1.82)  \\ 
  2 kernel ridge & 63.5(1.6) & 56.76(1.39) \\ 
  2 logistic & 74.98(5.98) & 70.18(9.18)  \\ 
  2 SVM & $61.08(1.41)^1$ & 56.72(2.62) \\ 
  2 RF & 62.26(1.35) & 56.18(1.16) \\ 
  causal\_rf 0.05 0 & 61.24(1.41) & 55.7(1.4) \\ 
  causal\_rf 0.01 0 & 61.16(1.38) & $55.06(1.38)^1$ \\ 
  causal\_rf 0.05 0.1 & 61.3(1.62) & 55.66(1.69)  \\ 
  causal\_rf 0.01 0.1 & $61.1(1.36)^2$ & 55.52(1.57) \\ 
   \hline
\end{tabular} \caption{The output table for a high dimensional data where a data point is equally likely to be assigned to the treatment group or control group.} \label{HighDTable}
\end{table}

For this data set, Table \ref{HighDTable} shows that our method ,without using matching, is highly competitive compared to the approaches that using the difference of two classification or regression methods as well as causal random forest approach.

%\subsection{Simulations where $\mu_T \neq \mu_C$}

%\textcolor{red}{I'm not convinced we need to do this, but it would be useful to do a sensitivity analysis on the previous type of result to misspecification of that ratio.}

%In practice, as the ratio of distributions $\frac{\mu_T}{\mu_C}$ are not known to us. Hence there is need to estimate it. We consider two methods to estimate the radon-nikodym derivatives. Note that the task of estimating the ratio is simpler than estimating the two distributions separately. We use both unconstrained least square importance fitting and Kullback-Leibler Importance Estimation procedure in estimating the ratio. 

%To be continued....

\subsubsection{Red Wine data set}

We provided experiment with the Red Wine data set in the main paper. We also perform similar experiment under different assignment mechanism settings for this data set.

\noindent \textbf{Setting 1: Equally likely to be assigned to be treatment or control group.}

\begin{table}[ht]
\centering\small
\begin{tabular}{rll}
  \hline
 & 0.01 & 0.1  \\ 
  \hline
linear causal SVM 1e-8 & 39.64(1.03) & 34.29(1.18)  \\ 
  linear causal SVM 1e-6 & 39.64(1.03) & 34.34(1.25) \\ 
  linear causal SVM 1e-4 & 39.62(1.07) & 34.34(1.22) \\ 
  quadratic causal SVM 1e-8 & 48.78(1.6) & 43.71(1.53) \\ 
  quadratic causal SVM 1e-6 & 48.74(1.59) & 43.75(1.74)  \\ 
  quadratic causal SVM 1e-4 & 49.08(2.53) & 43.65(2.57)  \\ 
  cubic causal SVM 1e-8 & 43.84(1.33) & 38.96(1.12)  \\ 
  cubic causal SVM 1e-6 & 40.2(1.28) & 35.39(1.16)  \\ 
  cubic causal SVM 1e-4 & 40.8(2.72) & 35.59(2.51)  \\ 
  rbf causal SVM 0.05,  1e-8 & 45.66(1.56) & 40.75(1.26)  \\ 
  rbf causal SVM 0.05, 1e-6 & 43.32(1.95) & 38.02(1.86)  \\ 
  rbf causal SVM 0.05, 1e-4 & $38.18(1.3)^2$ & $32.81(1.49)^2$ \\ 
  rbf causal SVM 0.1, 1e-8 & 45.99(1.87) & 40.81(1.65)  \\ 
  rbf causal SVM 0.1, 1e-6 & 44.51(1.45) & 39.51(1.33)  \\ 
  rbf causal SVM 0.1, 1e-4 & 39.19(1.78) & 33.85(1.44)  \\ 
  GenMatch, Ridge & 39.42(1.02) & 34.38(0.9)  \\ 
  Nearest, Ridge & 39.39(1.22) & 34.41(1.28)  \\ 
  Genmatch, kernel ridge & 39.1(1.63) & 34.31(1.59) \\ 
  Nearest, kernel ridge & 42.09(2.31) & 37.03(2.25)  \\ 
  2 ridge & 39.01(1.39) & 33.88(0.97) \\ 
  2 kernel ridge & 39.02(1.93) & 34.01(1.66) \\ 
  2 logistic & 39.08(1.12) & 33.88(0.88)  \\ 
  2 SVM & 42.85(1.75) & 37.6(1.73)  \\ 
  2 RF & $36.1(1.6)^1$ & $31.55(1.62)^1$ \\ 
  causal\_rf 0.05 0 & $38.31(1.01)^3$ & 33.22(1.06) \\ 
  causal\_rf 0.01 0 & 40.61(3.44) & 35.44(3.52)  \\ 
  causal\_rf 0.05 0.1 & 38.55(0.86) & $33.1(1.07)^3$  \\ 
  causal\_rf 0.01 0.1 & 41.91(3.2) & 36.81(3.19) \\ 
   \hline
\end{tabular} \caption{Output table for the red wine data set where each data pointequally likely to be assigned to be  treatment or control group.}\label{redwinetable}
\end{table}

For this setting,  from Table \ref{redwinetable},   the difference of two-random forest model seems to perform better than other algorithms. Causal SVM with RBF kernels performs similarly.

\noindent \textbf{Setting 2: The assignment mechanism is based on $\operatorname{Bernoulli}  \left( 0.75\left(\frac{1-\exp(-x_c^2)}{1+\exp(-x_c^2)}\right)\right)$}

In this setting, we let our assignment mechanism be depending on the reading of citric acid. We let $x_c$ denotes the reading of the citric acid and we let the probability that one is being assigned to the treatment group follows $\operatorname{Bernoulli}  \left( 0.75\left(\frac{1-\exp(-x_c^2)}{1+\exp(-x_c^2)}\right)\right)$.
\begin{table}[]
\centering\small
\begin{tabular}{rll}
  \hline
 & 0.01 & 0.1 \\ 
  \hline
linear causal SVM 1e-8 & 40.39(1.42) & 35.29(1.17)  \\ 
  linear causal SVM 1e-6 & 40.39(1.42) & 35.29(1.17)  \\ 
  linear causal SVM 1e-4 & 40.4(1.35) & 35.25(1.15) \\ 
  quadratic causal SVM 1e-8 & 48.02(2.32) & 42.9(2.43)  \\ 
  quadratic causal SVM 1e-6 & 47.92(2.09) & 42.78(2.28) \\ 
  quadratic causal SVM 1e-4 & 48.99(2.16) & 43.64(1.9)  \\ 
  cubic causal SVM 1e-8 & 44.52(1.28) & 39.41(1.35) \\ 
  cubic causal SVM 1e-6 & 41.7(1.36) & 36.54(1.42) \\ 
  cubic causal SVM 1e-4 & 44.25(3.07) & 38.61(2.65) \\ 
  rbf causal SVM 0.05,  1e-8 & 45.19(2.02) & 40.02(1.73)  \\ 
  rbf causal SVM 0.05, 1e-6 & 42.98(1.71) & 38.19(1.64)  \\ 
  rbf causal SVM 0.05, 1e-4 & $39.39(1.44)^2$ & $34.49(1.48)^2$  \\ 
  rbf causal SVM 0.1, 1e-8 & 45.41(1.27) & 40.24(1.54) \\ 
  rbf causal SVM 0.1, 1e-6 & 44.78(1.65) & 39.6(1.6)  \\ 
  rbf causal SVM 0.1, 1e-4 & 40.65(1.47) & 35.36(1.35)  \\ 
  GenMatch, Ridge & $40.11(1)^3$ & 34.96(0.98)  \\ 
  Nearest, Ridge & 41.41(1.43) & 36.12(1.52) \\ 
  Genmatch, kernel ridge & 41.3(1.33) & 36.3(1.21)  \\ 
  Nearest, kernel ridge & 43.15(1.4) & 37.84(1.26)  \\ 
  2 ridge & 40.32(1.04) & $34.92(1.01)^3$  \\ 
  2 kernel ridge & 40.49(1.64) & 35.29(1.53)  \\ 
  2 logistic & 40.34(0.84) & 35.11(1)  \\ 
  2 SVM & 43.02(2.21) & 37.71(2.05)  \\ 
  2 RF & $38.26(1.02)^1$ & $32.99(1.08)^1$ \\ 
  causal\_rf 0.05 0 & 41.45(1.41) & 36.19(1.38)  \\ 
  causal\_rf 0.01 0 & 50.98(4.76) & 44.55(3.53)  \\ 
  causal\_rf 0.05 0.1 & 41.48(1.45) & 36.18(1.62) \\ 
  causal\_rf 0.01 0.1 & 51.51(4.62) & 45.19(3.81) \\ 
   \hline
\end{tabular} \caption{ Output table for the red wine data where the assignment mechanism is based on $\operatorname{Bernoulli}  \left( 0.75\left(\frac{1-\exp(-x_c^2)}{1+\exp(-x_c^2)}\right)\right)$} \label{075table}
\end{table}
Table \ref{075table} shows that the 2-random forest method achieves the best performance, but the performance for most methods is very similar.

\noindent \textbf{Setting 3: The assignment mechanism is based on $\operatorname{Bernoulli}  \left( 0.5\left(\frac{1-\exp(-x_c^2)}{1+\exp(-x_c^2)}\right)\right)$}

In this setting, we let our assignment mechanism be depending on the reading of citric acid. We let $x_c$ denotes the reading of the citric acid and we let the probability that one is being assigned to the treatment group follows $\operatorname{Bernoulli}  \left( 0.5\left(\frac{1-\exp(-x_c^2)}{1+\exp(-x_c^2)}\right)\right)$.

\begin{table}[]
\centering\small
\begin{tabular}{rll}
  \hline
 & 0.01 & 0.1  \\ 
  \hline
linear causal SVM 1e-8 & 40.62(1.53) & 35.5(1.42) \\ 
  linear causal SVM 1e-6 & 40.62(1.53) & 35.51(1.39) \\ 
  linear causal SVM 1e-4 & 40.64(1.44) & 35.56(1.32)  \\ 
  quadratic causal SVM 1e-8 & 49.99(0.99) & 44.44(1.23)  \\ 
  quadratic causal SVM 1e-6 & 49.71(0.99) & 44.36(1.12)  \\ 
  quadratic causal SVM 1e-4 & 50.21(1.93) & 44.78(2.02) \\ 
  cubic causal SVM 1e-8 & 44.69(1.92) & 39.7(2)  \\ 
  cubic causal SVM 1e-6 & 42.46(1.39) & 37.08(1.16)\\ 
  rbf causal SVM 0.05,  1e-8 & 46.48(1.49) & 41.38(1.32)  \\ 
  rbf causal SVM 0.05, 1e-6 & 44.12(1.64) & 39.06(1.59)  \\ 
  rbf causal SVM 0.05, 1e-4 & $39.21(0.94)^2$ & $34.05(0.92)^2$  \\ 
  rbf causal SVM 0.1, 1e-8 & 46.42(1.22) & 41.61(1.61)  \\ 
  rbf causal SVM 0.1, 1e-6 & 45.66(1.39) & 40.59(1.37)  \\ 
  rbf causal SVM 0.1, 1e-4 & $39.86(0.84)^3$ & $34.74(0.86)^3$ \\ 
  GenMatch, Ridge & 40.89(1.05) & 35.41(0.61)  \\ 
  Nearest, Ridge & 41.59(1.63) & 36.21(1.56)  \\ 
  Genmatch, kernel ridge & 41.96(1.59) & 36.62(1.39)  \\ 
  Nearest, kernel ridge & 43.18(1.47) & 37.84(1.42) \\ 
  2 ridge & 40.48(0.82) & 35.19(0.56) \\ 
  2 kernel ridge & 41.59(1.18) & 36.12(1.33)\\ 
  2 logistic & 41.06(0.73) & 35.48(0.67)  \\ 
  2 SVM & 41.82(2.6) & 36.55(2.75)  \\ 
  2 RF & $38.26(1.11)^1$ & $33.4(1.18)^1$  \\ 
  causal\_rf 0.05 0 & 41.3(2.1) & 35.94(1.96)  \\ 
  causal\_rf 0.01 0 & 50.89(4.61) & 44.71(4.16)  \\ 
  causal\_rf 0.05 0.1 & 41.4(2.52) & 36.11(2.52)  \\ 
  causal\_rf 0.01 0.1 & 48.75(5.93) & 42.88(5.05)  \\ 
   \hline
\end{tabular} \caption{Output table for the red wine data where the assignment mechanism is based on $\operatorname{Bernoulli}  \left( 0.5\left(\frac{1-\exp(-x_c^2)}{1+\exp(-x_c^2)}\right)\right)$}\label{05table}
\end{table}

From Table \ref{05table}, the two-random forest method again outperform all algorithm, however, our algorithm achieve similar result using a simpler model.

\subsubsection{Summary of Experiments}

Methods based on differences of two predictive models often use a richer class of functions than methods using a single model. Our experiments tend to favor the more complex model classes, such as difference of 2-SVM, despite the fact that there is no real theoretical principle underlying the use of 2-SVM.  There are some advantages to using a single model, beyond the tighter bound on the 0-1 loss, and generalization bounds, in particular, better control over the complexity of the model, a single global optimization problem to solve with a guarantee of optimality, which is easier to troubleshoot and trust.

\end{document}